# Hybrid metaheuristics for architectural design optimization under small evaluation and wall-clock budgets

Luciano Ambrosini — LA Architecture & Computational Design Consulting, Naples, Italy — luciano.ambrosini@outlook.com — ORCID 0000-0003-1529-2694

**Abstract.** Architectural and structural design optimization in parametric software affords about a thousand evaluations of an expensive model for problems with tens of variables, a regime for which most metaheuristics were not designed. On the side of artificial intelligence, two hybrid metaheuristics are proposed: a covariance matrix adaptation evolution strategy (CMA-ES) whose sampling is enriched with differential-evolution trial vectors, and a particle swarm with success-history parameter adaptation, differential-evolution injection and local refinement. A budget-aware delay of population-size reduction in adaptive differential evolution is also proposed. Under a hard cap of 1,200 evaluations, all counted, the CMA-ES hybrid ranks first on seven shifted and rotated functions at 10 and 30 variables and beats CMA-ES on six of seven functions at 30 variables. Ablations attribute this gain to the injection and show that a local search pays off only when its result re-enters the population. The delayed schedule improves three adaptive differential-evolution variants in 40 of 51 comparisons, with no loss. On the side of engineering application, in Grasshopper the CMA-ES hybrid and plain CMA-ES are never significantly worse than the optimizers of Galapagos, SilverEye, Opossum and Tunny on travelling-salesman instances and on a built 85-variable free-form shell, where they end 48 % below the best external tool in 15 minutes. On a 22-variable building-performance problem with binary choices, differential evolution leads instead. Shifting and rotating the functions reverses the rank of Harris hawks optimization. Solvers should therefore be selected by problem structure and stopping time, not by a single ranking.

## 1. Introduction

Parametric design environments have made numerical optimization an ordinary step of architectural and structural design. In Grasshopper, the visual programming environment of Rhinoceros (McNeel & Associates), a designer can connect geometry to a finite-element solver or to a building-performance model and hand the resulting black box to an optimizer without writing code (Rutten, 2013; Wortmann et al., 2022). The evaluation behind each candidate design is expensive: a structural analysis of a free-form shell or an energy and daylight simulation takes from a fraction of a second to minutes, so practical runs stop after a few hundred to a few thousand evaluations (Waibel et al., 2019a; Wetter and Wright, 2004; Wortmann, 2019a). At the same time, design problems are rarely small. Facade, roof and shell parametrizations routinely involve tens of continuous variables, sometimes mixed with discrete choices. Architectural design optimization (ADO) therefore operates in a regime that the evolutionary computation literature studies less often: moderate to high dimensionality, a budget of roughly $10^3$ function evaluations (FE), and a black box whose cost is dominated by simulation rather than by the optimizer.

The algorithms available to designers in this regime were mostly not built for it. Generic genetic algorithms remain the default in parametric software, although benchmarks on building and structural problems have repeatedly shown them to be among the weakest options at small budgets (Cichocka et al., 2017; Waibel et al., 2019a; Wortmann, 2019a). Model-based methods such as RBFOpt perform well when evaluations are very expensive and the dimension is low, but their own overhead grows quickly with the number of variables (Costa and Nannicini, 2018; Wortmann, 2017). The adaptive differential evolution and covariance-matrix-adaptation variants that win the CEC competitions are tuned for budgets of $10^4$ to $10^6$ evaluations (Brest et al., 2017; Tanabe and Fukunaga, 2014). Their population-size schedules, step-size settings and local-search phases assume that the run is long. Hybrids that combine covariance matrix adaptation with differential evolution, or that interleave a local search, exist (Ghosh et al., 2012; Kämpf and Robinson, 2009; Molina et al., 2018, 2010), but they have been designed and assessed at the same large budgets. How these mechanisms behave when the whole run is worth a thousand evaluations, and whether their extra evaluations pay for themselves, is largely unknown.

The way such algorithms are compared adds a second difficulty. Comparisons in design software often report best runs or a handful of repetitions, mix evaluation-based and wall-clock budgets, and let some solvers spend evaluations that are not counted against them. Synthetic test suites whose optimum lies at the centre of the search box can reward methods that are biased towards the centre rather than methods that search well (Kůdela, 2023, 2022). The proliferation of metaphor-based algorithms makes it harder still to tell which mechanism is responsible for a result (Campelo and Aranha, 2021; Sörensen, 2015). For a designer who must choose a solver for a specific problem, these confounds matter more than small differences in average rank.

This paper addresses the low-budget ADO regime from both sides: algorithm design and evaluation protocol. It presents two hybrid metaheuristics built for budgets of about $10^3$ evaluations and a budget-aware schedule for the L-SHADE family, and evaluates them, together with seven established algorithms, a portfolio selector and the optimizers of four Grasshopper tools, on four classes of problems that a designer actually encounters. All solvers are implemented in one Grasshopper plug-in, referred to here as SilverHawk, and in a command-line console that shares its algorithm code. The contributions are the following.

First, two algorithms are described: MERLIN, which interleaves differential-evolution trial vectors into the sampling step of CMA-ES and adds an inline Solis–Wets refinement, and RAPTOR, which combines particle-swarm dynamics with success-history parameter adaptation, Lévy-flight perturbation and differential-evolution injection. The paper also describes a delay cap on the linear population-size reduction of L-SHADE that keeps the population from collapsing before a short run has explored the space. All mechanisms are given in standard optimization terminology. The design principle that links them is methodological: components are interleaved within each generation instead of run in separate phases, and each component must repay the evaluations it consumes under the hard cap, which Section 6 tests by removal. To the author's knowledge, none of the optimizers documented for Grasshopper offers a covariance-matrix-adaptation search enriched with differential-evolution trial vectors, or a population-size schedule defined for budgets of about $10^3$ evaluations (Section 2.1), and none has been compared with the others under one stopping rule on problems that range from combinatorial to finite-element.

Second, the evaluation protocol is designed to remove the confounds listed above. Every solver is stopped at a hard cap on real evaluations that includes local-search and injection evaluations. The synthetic suite is run both in its original form and with random shifts and rotations. The finite-element problem is run under an equal wall-clock budget, so that optimizers with heavy internal computation pay for it. All comparisons use 30 independent runs on the synthetic and surrogate problems and non-parametric tests (Carrasco et al., 2020; Derrac et al., 2011).

Third, the evidence covers four problem classes: eight synthetic functions at D = 10 and D = 30 in original and shifted-rotated form; travelling-salesman instances encoded through integer sliders, solved inside Grasshopper against the tools designers use; three single-objective formulations of a published 22-variable building-performance problem with binary design choices (Zorn et al., 2025); and the 85-variable free-form concrete shell derived from the Meiso no Mori crematorium (Toyo Ito, Kakamigahara, Japan, 2006), a structural benchmark with a long history in computational design (Cichocka et al., 2017; Pugnale and Sassone, 2007). Figure 1 summarizes the design of the study. The results show that the preferred solver depends on the structure of the problem and on the budget horizon rather than on a single ranking. Against the optimizers of the Grasshopper tools, the two best solvers of the plug-in, its CMA-ES implementation and MERLIN, are never significantly worse on any problem, better than the genetic and particle-swarm tools on every instance, and on the shell they end 48 % below the best external tool in the same time, a difference that does not depend on excluding the inert inputs of the model. The results also show that the centre bias of the original synthetic suite reverses at least one widely repeated conclusion.

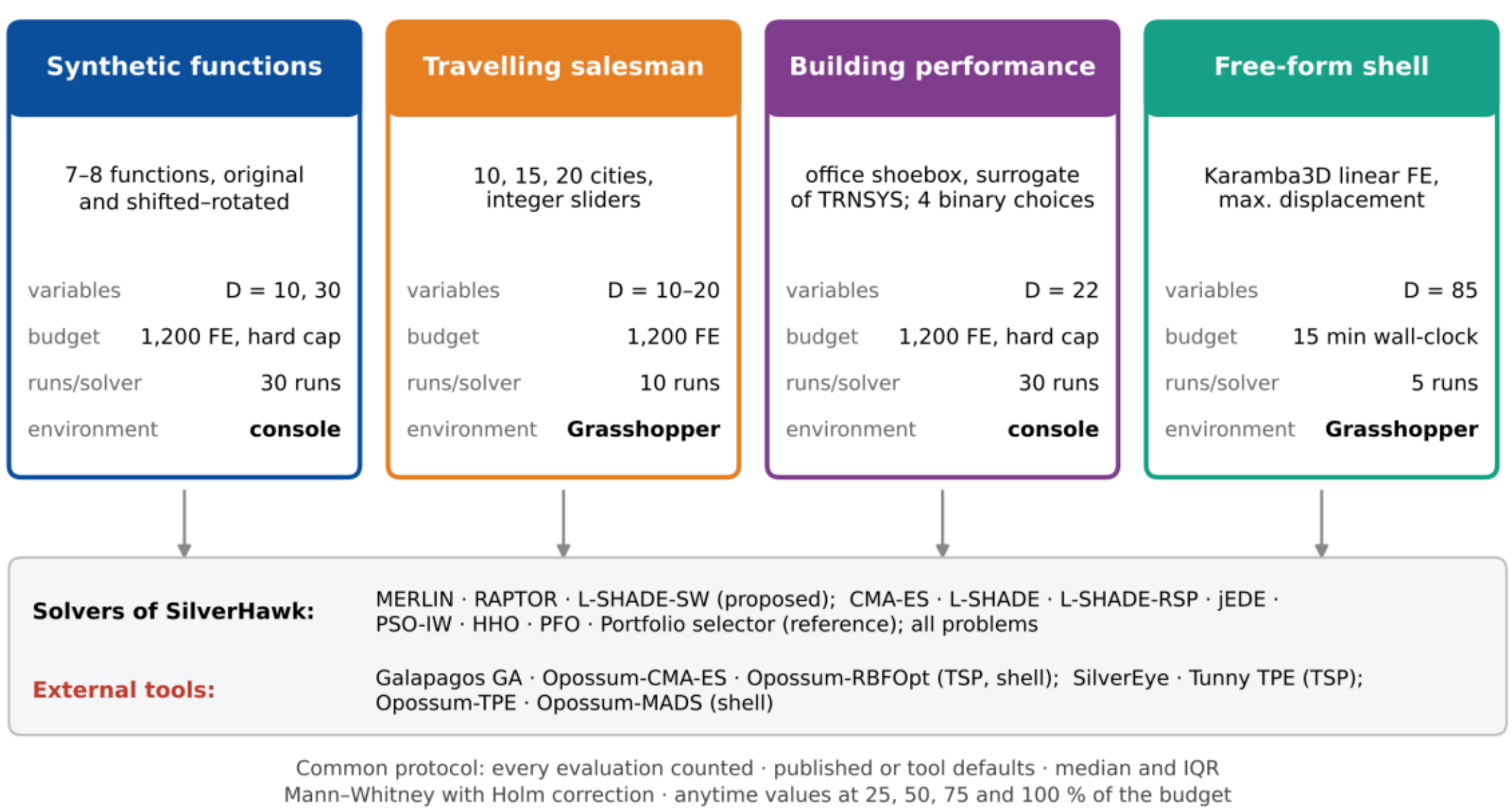

**Figure 1.** Design of the study. Four problem classes differ in structure, cost per evaluation and environment; all are solved by the same roster under a common protocol. External Grasshopper tools are run only on the problems solved inside Grasshopper.

Fourth, ablation studies under the same hard cap identify which components carry the performance. The differential-evolution injection is the decisive part of MERLIN on rotated landscapes, while its Solis–Wets refinement does not repay its evaluations at D = 30. The delay cap improves all three L-SHADE variants, while a comparable horizon rescaling for Harris hawks optimization does not.

Section 2 positions the work with respect to optimization in parametric design, low-budget metaheuristics, hybrid and memetic algorithms, and benchmarking practice. Section 3 describes the algorithms. Section 4 details the problems and the experimental protocol. Section 5 reports the results and Section 6 the ablations. Section 7 discusses solver selection, the budget horizon and the limitations of the study, and Section 8 concludes.

# 2. Related work

## 2.1 Optimization inside parametric design environments

Grasshopper offers several optimization tools, each built around one algorithmic family. Galapagos provides a genetic algorithm and simulated annealing (Rutten, 2013); Octopus and Wallacei provide multi-objective evolutionary algorithms (Makki et al., 2018; Vierlinger, 2013); Opossum wraps the radial-basis-function method RBFOpt and, in later versions, CMA-ES and other population-based methods (Costa and Nannicini, 2018; Wortmann, 2017); SilverEye implements particle swarm optimization (Cichocka et al., 2017); Optimus implements a self-adaptive differential evolution with an ensemble of mutation strategies (Cubukcuoglu et al., 2019); and Tunny exposes the samplers of the Optuna library (Natsume, 2022). In the versions described in the cited sources, none of these tools combines a covariance-matrix-adaptation search with differential-evolution trial vectors counted against the budget, or offers a population-size schedule calibrated for budgets of about $10^3$ evaluations. Surveys of practitioners show that most users rely on whichever tool they know, usually a genetic algorithm, and run it with default settings (Wortmann et al., 2022; Wortmann, 2019b).

Benchmarks in this setting consistently point in the same direction. On building-energy and structural problems with small budgets, model-based methods and well-adapted evolutionary or swarm methods outperform the default genetic algorithm (Waibel et al., 2019a; Wortmann, 2019a; Wortmann et al., 2015). The Kakamigahara shell, introduced as a morphogenesis problem by Pugnale and Sassone (2007) and later used to compare SilverEye, Galapagos and Octopus (Cichocka et al., 2017), is one of the few structural problems that has been reused across studies. These benchmarks, however, compare tools rather than mechanisms. Each tool ships one family with its own defaults, and the comparisons cannot separate the effect of the algorithm from that of its implementation, its stopping rule or its overhead. Wortmann and Waibel argued for a common test bed for this reason (Waibel et al., 2019b). The present study keeps the tools as external baselines but compares mechanisms inside a single implementation, under one stopping rule.

## 2.2 Metaheuristics at small evaluation budgets

The performance of a metaheuristic depends strongly on the budget at which it is assessed. Rankings obtained at the budgets of the CEC competitions can change substantially when the number of evaluations is reduced by one or two orders of magnitude (Kazíková et al., 2021; Piotrowski et al., 2025). Successful adaptive differential evolution variants such as SHADE, L-SHADE, jSO and L-SHADE-RSP adapt their control parameters from success history and shrink their population linearly over the run (Brest et al., 2017; Stanovov et al., 2018; Tanabe and Fukunaga, 2014, 2013). The reduction schedule is defined as a function of the evaluations already spent, so a short run reaches the minimum population size after only a few generations. CMA-ES, the reference method for non-separable continuous problems, has to learn a covariance matrix with D(D+1)/2 entries (Hansen et al., 2003); at D = 30 and above, a budget of $10^3$ evaluations is too short for that estimate to mature. Swarm methods such as PSO and Harris hawks optimization converge quickly in early generations but are sensitive to velocity limits and to the scaling of their exploration phase (Clerc and Kennedy, 2002; Heidari et al., 2019).

Surrogate-assisted methods address small budgets directly by modelling the objective (Costa and Nannicini, 2018; Wortmann et al., 2015). Their cost per iteration grows with the number of samples and variables, which limits them in the high-dimensional part of the ADO regime and on problems whose evaluation time is comparable with model fitting. This paper stays within model-free metaheuristics and asks how their mechanisms must be adapted when the whole run is about $10^3$ evaluations.

## 2.3 Hybrid and memetic algorithms

Combining covariance matrix adaptation with differential evolution is not new. Differential covariance matrix adaptation (Ghosh et al., 2012), the hybrid CMA-ES/HDE of Kämpf and Robinson, applied to the solar potential of urban forms (Kämpf and Robinson, 2009), and redistribution of CMA-ES individuals with differential evolution (Chen and Liu, 2022) all exploit the complementary strengths of the two families. LSHADE-SPACMA combines L-SHADE with CMA-ES in one run and was assessed on the CEC 2017 suite at budgets of $10^4$ D evaluations (Mohamed et al., 2017); MERLIN differs in keeping a single CMA-ES search distribution, to which differential-evolution vectors are added as extra offspring that take part in its update, and in counting every evaluation at a budget of about $10^3$. Modular frameworks have explored PSO-DE hybrids systematically (Boks et al., 2020). Memetic algorithms add a local search to a global method: MA-SW-Chains and SHADE-ILS apply Solis–Wets or other local searches in dedicated phases and won large-scale CEC competitions (Molina et al., 2018, 2010; Solis and Wets, 1981). These designs share two assumptions that do not hold in ADO. They allocate a separate phase or subpopulation to each component, which is affordable only when the budget is large. They also count local-search evaluations against budgets of $10^5$ or more, where a few extra evaluations per generation are negligible. At $10^3$ evaluations, every evaluation spent by a component must be justified by its effect on the result. The algorithms in Section 3 interleave their components within each generation instead of alternating phases, and Section 6 tests whether each component earns its evaluations under a hard cap.

### 2.4 Benchmarking practice and algorithm selection

The no-free-lunch theorems state that no algorithm dominates over all problems (Wolpert and Macready, 1997). In practice this places the burden on the design of comparisons. Best-practice guidelines also warn that comparisons at equal numbers of evaluations can hide differences in the computational cost of each evaluation (Bartz-Beielstein et al., 2020; Sala and Müller, 2020), and a recent position paper argues for wall-clock time as the primary budget (Lian, 2025). Recommended practice includes multiple independent runs, non-parametric tests with corrections for multiple comparisons, and reporting of distributions rather than best values (Carrasco et al., 2020; Derrac et al., 2011; Sala and Müller, 2020). Two further issues have received attention recently. First, many algorithms published as new are re-descriptions of known mechanisms under a natural metaphor, which obscures what actually drives performance (Campelo and Aranha, 2021; Sörensen, 2015). Second, algorithms with a structural bias towards the centre of the search space look strong on test functions whose optimum is at the origin and lose that advantage once the functions are shifted (Kůdela, 2023, 2022). Both issues concern algorithms that are widely used in architectural design studies, where synthetic functions with centred optima remain the usual sanity check.

When no single algorithm dominates, the practical question becomes which algorithm to use on a given problem. Per-instance algorithm selection based on landscape features addresses this question when features can be computed cheaply relative to the budget (Bischl et al., 2016; Kerschke and Trautmann, 2019). In ADO, feature sampling competes with the optimization itself for the same few hundred evaluations. The present study does not propose a selector. It provides the empirical basis for a simple selection rule organised by problem structure and budget horizon, and it reports one portfolio selector as a reference point.

## 3. Algorithms

All algorithms in this study operate in the normalized box [0, 1]^D and map candidates to the design variables only for evaluation. They are implemented as ask–tell state machines: the host asks for one candidate, evaluates it, and returns its objective value before the next candidate is requested. This interface is imposed by the parametric environment, where every evaluation means recomputing the geometric and analysis pipeline of the definition, and it has one consequence that shapes the designs below. Any evaluation an algorithm requests, whether it comes from the main population, from an injected vector or from a local-search probe, is a real evaluation of the black box and is counted against the same budget. Section 4 describes how this rule is enforced in the experiments.

The names MERLIN and RAPTOR are labels only: no mechanism is derived from a natural metaphor, and each algorithm is defined below by its components (Sörensen, 2015). Three design principles follow from the ADO regime introduced in Section 1. First, components are interleaved within each generation rather than run in separate phases, because a budget of about $10^3$ evaluations leaves no room for a dedicated exploration phase followed by a dedicated refinement phase. Second, each component must earn the evaluations it consumes; this is tested in Section 6 by removing components under a hard cap. Third, schedules that depend on the total budget, such as population-size reduction, must be calibrated for small budgets, and they can only be applied when the total budget is known before the run starts.

### 3.1 MERLIN: CMA-ES with interleaved differential-evolution injection

MERLIN keeps the sampling and update equations of CMA-ES (Hansen et al., 2003) and changes what is sampled. In each generation, $\lambda - K$ of the $\lambda$ offspring are drawn from the search distribution $N(m, \sigma^2 C)$ as in standard CMA-ES. The remaining K offspring are differential-evolution trial vectors built from an archive of good solutions found so far. All $\lambda$ offspring are evaluated, ranked together and used together in the update of m, C and $\sigma$. Figure 3 shows one generation. The injected vectors act as external proposals that the distribution can follow when they are better than its own samples, a mechanism that Hansen showed to be compatible with the CMA-ES update provided that the injected points enter the recombination like any other selected point (Hansen, 2011).

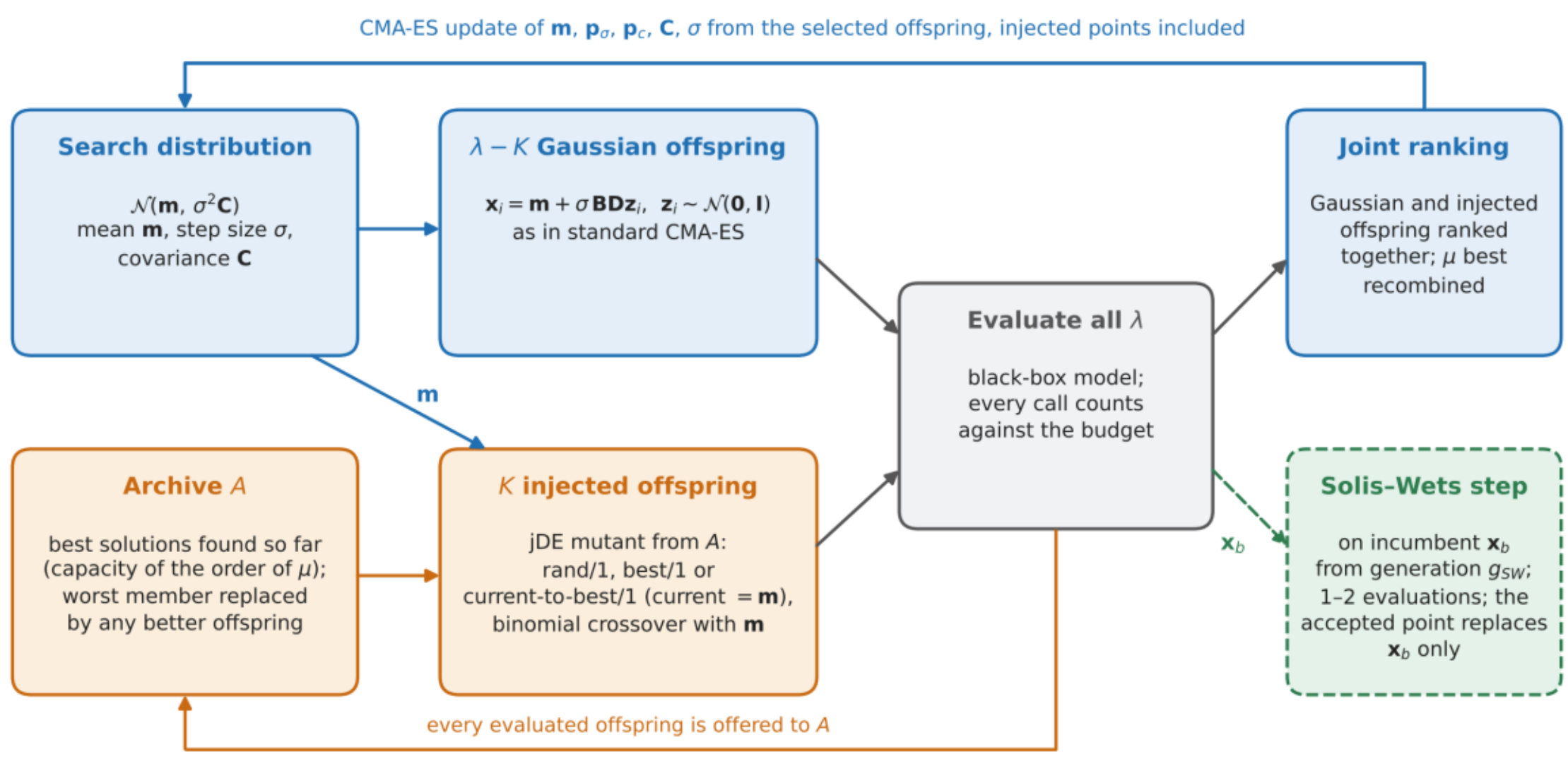

**Figure 3.** One generation of MERLIN. $\lambda - K$ offspring are sampled from the CMA-ES distribution and K are differential-evolution trial vectors built from an archive; all $\lambda$ are evaluated and ranked together, and the CMA-ES update uses the selected offspring of both kinds. The Solis–Wets step (dashed) acts on the incumbent only.

The population size follows the usual CMA-ES default unless the user sets a larger one, and the number of injected offspring is a fixed fraction $\rho = 0.2$ of $\lambda$. The distribution starts at the centre of the box with an initial step size $\sigma_0 = 0.1$ in normalized coordinates, smaller than the value of 0.3 used for the CMA-ES baseline. Both values are tested in Section 6. The injected vectors use standard differential-evolution mutation and crossover with self-adapted control parameters (Brest et al., 2006). From a fixed generation onward, MERLIN also performs a Solis–Wets refinement step (Solis and Wets, 1981) around the incumbent best, which costs one or two evaluations per generation. An accepted refinement point replaces the incumbent but enters neither the archive nor the CMA-ES update; this detail matters for the ablation in Section 6. Further implementation details are available from the author on request.

## 3.2 RAPTOR: particle swarm with success-history adaptation, Lévy relocation and DE injection

RAPTOR starts from the global-best particle swarm (Clerc and Kennedy, 2002; Eberhart and Kennedy, 1995) and adds four mechanisms, each addressing a known weakness of PSO at small budgets. The inertia weight and the acceleration coefficients are adapted per particle from a success history, in the manner of SHADE (Tanabe and Fukunaga, 2013). Particles whose personal best stagnates are relocated around the global best by Lévy-distributed steps (Mantegna, 1994). After each move, the worst particles are replaced by differential-evolution trial vectors built from an archive, as in MERLIN. Finally, a Solis–Wets step is performed around the global best; unlike in MERLIN, an accepted point is written into the swarm and takes part in the following generations. Velocities are clamped to a fraction $v_{max} = 0.5$ of the variable range, a value tested in Section 6.

## 3.3 L-SHADE-SW and a budget-aware population-size schedule

L-SHADE-SW is L-SHADE (Tanabe and Fukunaga, 2014) with the same interleaved Solis–Wets step: from a fixed generation onward, one step is performed around the elite solution, and an accepted point replaces the worst member of the population. The design is the low-budget counterpart of SHADE-ILS (Molina et al., 2018), which alternates differential evolution and local search in separate phases. All other components are those of L-SHADE.

The schedule that this paper proposes for the L-SHADE family concerns the linear population-size reduction (LPSR). In L-SHADE the population shrinks linearly with the evaluations spent, from $N_{init}$ to $N_{min}$, starting in the first generation. With $N_{init} = 20$ and a budget of 1,200 evaluations, the population has lost a quarter of its members after 300 evaluations, when a 30-dimensional landscape has barely been sampled. The onset of the reduction is delayed by a fraction $\delta$ of the budget. The delay is zero for budgets above a reference budget, grows as the budget becomes smaller, and is capped so that some reduction always takes place at the end of a short run; after the delay, the reduction is compressed linearly into the remaining part of the run. At $FE_{max} = 1{,}200$ the delay is $\delta = 0.76$: the population keeps its 20 members for three quarters of the run and shrinks to $N_{min}$ in the last quarter. Figure 2 shows the resulting trajectories for three budgets. The schedule adds no evaluations and no parameters to adapt during the run, and it reduces to standard LPSR for large budgets. It is applied identically to L-SHADE, to L-SHADE-SW and to an implementation of the rank-based selective pressure variant L-SHADE-RSP (Stanovov et al., 2018).

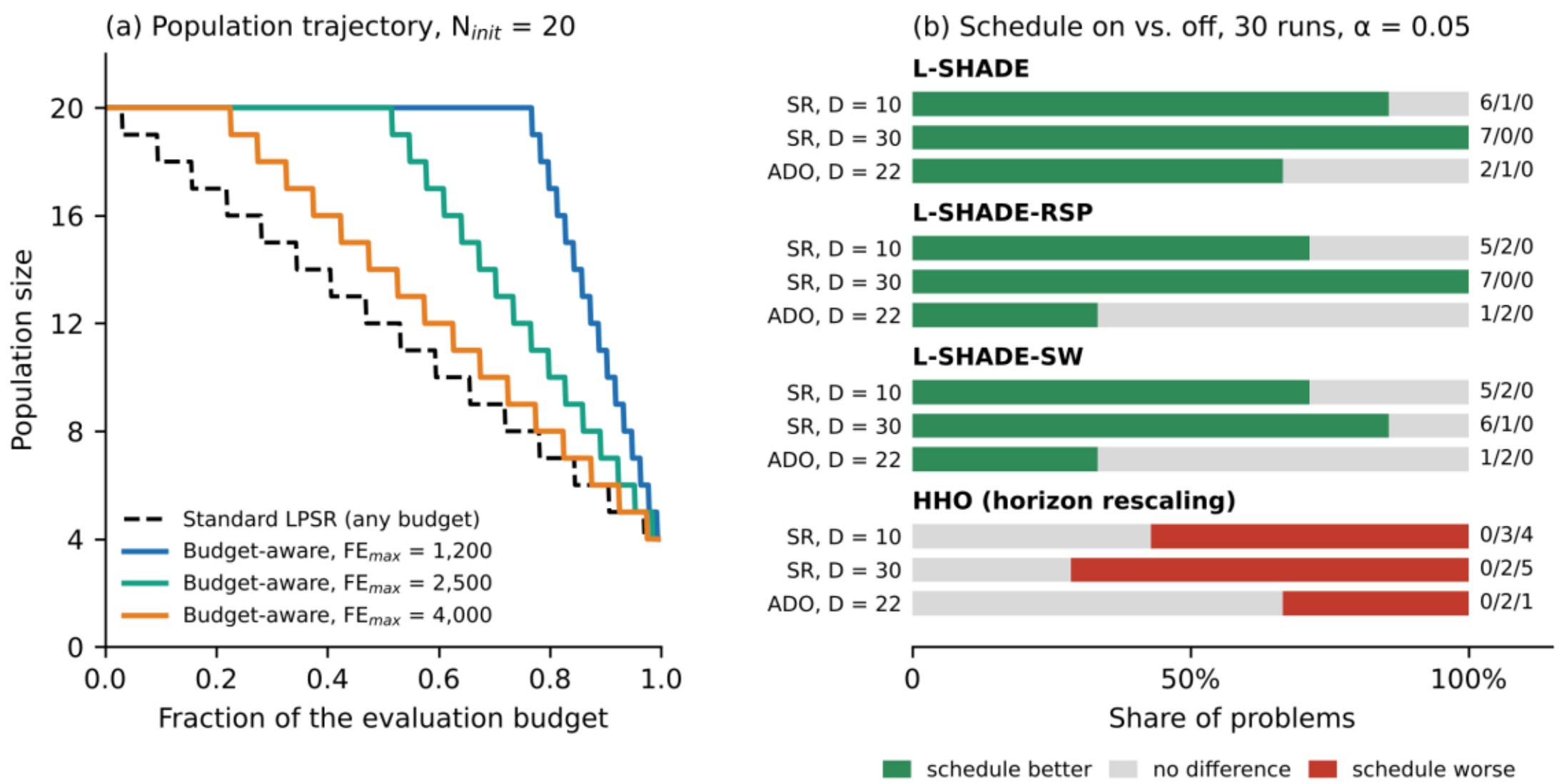

**Figure 2.** Budget-aware population-size schedule. (a) Population size of the L-SHADE variants over the evaluation budget, standard linear reduction and budget-aware delay for three budgets. (b) Wins, ties and losses of the schedule against the standard reduction on the shifted–rotated functions (SR) at D = 10 and D = 30 and on the building problem (ADO, three objectives), 30 runs, two-sided Mann–Whitney, $\alpha = 0.05$; bottom: the analogous horizon rescaling of HHO.

The schedule requires FE_max to be known when the run starts. This holds under a hard evaluation cap, but not when a run is stopped by wall-clock time, as in the shell problem (Section 4.5). Section 3.5 describes what happens in that case.

The same idea was applied to Harris hawks optimization (Heidari et al., 2019), whose escaping energy decides between exploration and exploitation and decays over the run: stretching its horizon in short runs keeps the hawks exploring for longer. Section 6 shows that this rescaling degrades HHO under a hard cap, and it is reported as a negative result rather than as part of the contribution.

### 3.4 Portfolio selector

SilverHawk also contains a landscape-aware portfolio selector that chooses among several of the algorithms above during a run. Its mechanism is outside the scope of this paper. It is included in the tables as a reference row, run with its default settings, so that the reader can compare the individual solvers with an automatic choice made under the same budget.

### 3.5 Execution in the parametric host

The console used for the synthetic and surrogate experiments queries an algorithm until it returns no further candidate, so every component described above is executed. Inside SilverHawk, the host loop requests at most N candidates per generation, N being the population size set by the user, and then advances the generation. Two consequences follow for the Grasshopper experiments (Sections 4.3 and 4.5).

First, a refinement step placed after the main offspring is reached only when the main offspring are fewer than N. In MERLIN and RAPTOR the main offspring number exactly N at the dimensions of this study ($\lambda = N$ since $4 + \lfloor 3 \ln D \rfloor < N$), so the Solis–Wets step is never executed and the Grasshopper results correspond to the configurations without Solis–Wets that are ablated in Section 6. In L-SHADE-SW the step is executed only once the population has started to shrink.

Second, when the run is stopped by wall-clock time, the host sets the nominal number of iterations to a very large value, so that FE_max is effectively infinite. For the L-SHADE variants this gives $\delta = 0$ and FE/FE_max close to zero for the whole run: the population never shrinks and these solvers run with a constant population of N individuals. The same setting removes the decay of the escaping energy in HHO, which keeps E close to $2E_0$ throughout the run. These are properties of the host rather than of the algorithms, and they are declared wherever the Grasshopper results are discussed.

# 4. Experimental design

The study uses four classes of problems that differ in structure, in cost per evaluation and in the environment in which they are solved (Table 1). The synthetic functions and the building-performance problem are solved in a console that hosts the same algorithm implementations as SilverHawk and allows thousands of runs under a hard evaluation cap. The travelling-salesman instances and the shell are solved inside Grasshopper, where every evaluation recomputes the parametric definition, so that the comparison includes the optimization tools a designer would actually use.

## 4.1 Budgets and stopping rules

All console runs stop at a hard cap of 1,200 evaluations, counted on the black box itself. Evaluations spent by injected vectors and by Solis–Wets probes count against the cap like any other. A solver whose generation would exceed the cap is stopped mid-generation, and a solver that requests fewer evaluations (the L-SHADE variants, whose population shrinks) is not compensated. The nominal horizon passed to each algorithm is $\lceil 1{,}200 / N \rceil$ generations, so that schedules that depend on the budget see the same horizon in every run. This rule matters because a hybrid that spends two extra evaluations per generation gains about 10 % more evaluations over a run of this length if extra evaluations are not counted. The population size is N = 20 for all population-based solvers, the default of SilverHawk and a common choice in architectural optimization studies (Cichocka et al., 2017; Wortmann, 2017).

The travelling-salesman runs in Grasshopper use the same budget of 60 generations of 20 candidates (1,200 evaluations). The shell problem uses a wall-clock budget of 15 minutes per run, which corresponds to 2,000–2,450 evaluations of the finite-element model on the test machine. A time budget is the natural stopping rule for a designer and it charges each solver for its own computational overhead. Each solver receives a single run in that time; restarts, which a fixed-time protocol may allow (Lian, 2025), are not used. It also makes the number of evaluations depend on the solver, which is why the shell results are analysed separately and why Section 3.5 describes how the host behaves in this mode.

Each console configuration is run 30 times with seeds 5000 + 137r, r = 0, …, 29, shared by all solvers. Travelling-salesman configurations are run 10 times, and the shell 5 times per solver.

## 4.2 Synthetic functions in original and shifted–rotated form

Eight functions cover separable and non-separable, unimodal and multimodal landscapes: Sphere, Rosenbrock, Rastrigin, Ackley, Griewank, Schwefel 2.26, Schwefel 1.2 and Step, at D = 10 and D = 30, with their usual search domains (Jamil and Yang, 2013). In their usual form (set F1) all but Rosenbrock and Schwefel 2.26 have the optimum at the centre of the box, and an algorithm that is biased towards the centre gains an advantage that has nothing to do with search ability (Kůdela, 2022).

The second set (F2) applies a shift and a rotation to the same functions, f_SR(x) = f(M(x − o) + x*), where x* is the optimum of the original function, o is drawn uniformly within the central 80 % of the box in each coordinate, and M is a random orthogonal matrix obtained by Gram–Schmidt orthogonalization of a Gaussian matrix. The new optimum is at x = o with the original optimal value. Schwefel 2.26 is left out of F2: its definition relies on the box, and after a rotation points inside the box map to arguments outside the original domain, where the function takes values below its nominal optimum. F2 therefore contains seven functions. The CEC and BBOB suites are not used because they are defined for budgets of $10^4$ D evaluations or more, and because the comparison of F1 with F2 needs two sets that differ only by the transformation; running the algorithms on those suites at about $10^3$ evaluations is left for future work. Shift vectors and rotation matrices are generated from a fixed seed for each function and dimension and are distributed with the supplementary material, so every solver sees the same instances.

## 4.3 Travelling-salesman instances in Grasshopper

Three symmetric Euclidean instances with 10, 15 and 20 cities were generated for this study as uniform random points in a 10 × 10 square (Grasshopper Populate 2D). The coordinates are fixed and distributed with the supplementary material, so that every solver and every reconstruction of the tours refers to the same geometry. The Grasshopper definition is the travelling-salesman exercise distributed with SilverEye (Cichocka et al., 2017), with these city sets in place of the original ones: each city has an integer slider in [0, N − 1] that assigns it to a branch of a data tree, cities in the same branch are visited in index order, and the branches are flattened in ascending order into a closed polyline whose length is the objective. The mapping from sliders to tours is many-to-one: whole regions of the slider space give the same tour, so the landscape consists of plateaus separated by steps. This structure differs sharply from the synthetic functions and resembles design problems with ordering or assignment decisions, which designers routinely express through integer sliders. The optimum of each instance was computed exactly with the Held–Karp dynamic-programming algorithm (Held and Karp, 1962): 39.06, 44.78 and 48.46 for 10, 15 and 20 cities. The first two were reached in the present runs; the 20-city optimum was not reached by any solver. Because the many-to-one encoding can favour some solvers, a control experiment on the same 10- and 15-city sets uses a bijective random-key encoding (Bean, 1994), in which the tour visits the cities in the order of continuous slider values; it is reported in the supplementary material.

The instances are solved inside Grasshopper by the solvers of SilverHawk and by four external tools: Galapagos (genetic algorithm), Opossum (RBFOpt and CMA-ES), SilverEye (particle swarm) and Tunny (tree-structured Parzen estimator, run three times only because a single run on the 20-city instance takes over an hour). All receive 1,200 evaluations: 60 generations of 20 candidates for the population-based solvers, an evaluation limit of 1,200 for Opossum and Tunny, and, for Galapagos, the best value reached within the first 1,200 evaluations of a longer run. External tools are used with the defaults of the version installed at the time of the campaign, unchanged in this study; SilverEye runs only in Rhinoceros 5 and was run there. The SilverHawk runs also had a stagnation stop, set between 50 generations and no limit depending on the batch; it ended one run out of 330 early (Harris hawks optimization on the 20-city instance, after 56 generations and 1,120 evaluations), and all other runs used the full 60 generations. The L-SHADE variants used 1,113–1,132 evaluations because of their population reduction.

## 4.4 Building-performance problem

The building-performance problem is the shoebox office model published by Zorn et al. (2025) together with a dataset of TRNSYS simulations. It has 22 variables: fifteen coordinates that define the geometry of five rows of shading slats, the window sill height, the maximum air-change rate, the insulation thickness, and four binary construction choices. The three objectives are the annual energy demand $Q_{tot}$ (kWh $m^{-2}$ $a^{-1}$), the overheating hours $K_h$ (h) and the daylight factor DF, which is maximized. Each objective is solved separately as a single-objective problem, which gives three problems with the same variables and different landscapes.

The objectives are evaluated by surrogate models trained on the public dataset, following the model classes that Zorn et al. found best (gradient-boosted regression for $Q_{tot}$ and DF, a multilayer perceptron for $K_h$), but trained on the full set of about $10^4$ samples. Their cross-validated coefficient of determination is at least 0.98 for all three targets. Variable bounds follow Zorn et al. and are enforced as hard limits, so that every candidate lies within the region covered by the training data. The surrogate makes 30 runs per solver affordable; its accuracy outside that region is irrelevant here because no candidate leaves it. The binary variables are handled as continuous variables rounded at evaluation, as a designer would do with a slider.

## 4.5 Free-form shell: the Meiso no Mori crematorium

The structural problem is the concrete shell roof of the Meiso no Mori crematorium in Kakamigahara (Toyo Ito & Associates, 2006; structural design by Mutsuro Sasaki), a free-form surface introduced as an optimization benchmark by Pugnale and Sassone (2007) and later used to compare particle swarm, genetic and evolutionary solvers in Grasshopper (Cichocka et al., 2017). The surface is controlled by the vertical position of 100 control points, each in [−5, +5] m. Fifteen of them are fixed by the supports: the definition overwrites their values, so they do not affect the objective, and 85 variables are effective. The objective is the maximum displacement of the shell under self-weight, computed for a 20 cm concrete shell by a linear finite-element analysis in Karamba3D (Preisinger, 2013). One evaluation takes about 0.4 s on the test machine. With all variables at zero the surface is flat, and its maximum displacement is 85.4 cm.

Sixteen solvers were run: eleven from SilverHawk and five external ones, namely Galapagos, Opossum-RBFOpt, Opossum-CMA-ES and the two solvers that Opossum added in version 3.2, TPE (tree-structured Parzen estimator) and MADS (mesh adaptive direct search). The SilverHawk solvers use a 15-minute limit and stop earlier only if the best value does not improve for 100 generations. They receive the 85 effective variables, because the host lets the user exclude fixed inputs from the search; the external tools receive all 100 sliders, as a designer would connect them. Section 6.4 tests whether this difference matters. Galapagos is stopped manually at 15 minutes. The two Opossum solvers are given the number of evaluations that fit into 15 minutes in a preliminary run (418 for RBFOpt, whose model fitting dominates the cost, and 2,370 for CMA-ES). Their wall-clock cost per evaluation, measured on the campaign runs, is 0.38 s for Galapagos and Opossum-CMA-ES and 2.15 s for Opossum-RBFOpt.

TPE and MADS were run after the main campaign, with Opossum 3.2.1, default solver settings (only the number of evaluations was user-set; the solver-specific settings are internal to the tool and not public) and all 100 sliders, on the same definition and machine (15 October 2026). Opossum has no default time-limit setting, so their number of evaluations was preset from a preliminary run (1,130 for TPE and 1,900 for MADS). Their logs time-stamp every evaluation with a resolution of one second, which allows each run to be cut exactly: the final value of a run is the best value found within 900 s of the first logged evaluation, and later evaluations are discarded. Within that time TPE completed 1,046–1,100 evaluations and MADS 1,354–1,713. The cost per evaluation of MADS is constant (0.5–0.7 s), whereas that of TPE grows from 0.56 s over the first hundred evaluations to about 1.15 s over the last hundred, as its model is refitted to a growing set of trials.

Two conditions of the host were controlled because they change the number of evaluations that fit into the time limit. The Rhinoceros viewport was not redrawn during the runs of SilverHawk (redrawing the shell after each evaluation lowers the throughput from about 2.6 to 1.8 evaluations per second), and every run started from a freshly loaded model; it was verified for each run that the first generation evaluates new candidates (first-generation best between 3.3 and 7.0 cm) and that the best value improves within the first ten generations. Under these conditions the SilverHawk solvers performed between 1,955 and 2,457 evaluations per run. Their cost per evaluation, 0.37–0.46 s, matches that of Galapagos and Opossum-CMA-ES, so the external tools did not run under a heavier display load. One solver, PFO, is an exception: after about ten generations its population collapses onto a few points, the definition is not recomputed for unchanged inputs, and the run stops by stagnation after one to three minutes. Its later evaluations are repetitions, so PFO is reported without being treated as time-fair.

## 4.6 Solvers

Table 2 lists the solvers. Besides the three algorithms of Section 3, the roster contains the reference methods of each family with their published settings: CMA-ES (Hansen et al., 2003) with $\sigma_0$ = 0.3 in normalized coordinates, L-SHADE (Tanabe and Fukunaga, 2014), an implementation of the rank-based selective pressure of L-SHADE-RSP (Stanovov et al., 2018) on the same L-SHADE base without the features designed for large budgets, the self-adaptive differential evolution with an ensemble of mutation strategies (jEDE) that underlies the Optimus plug-in (Brest et al., 2006; Cubukcuoglu et al., 2019), particle swarm with inertia weight (PSO-IW) (Shi and Eberhart, 1998) with the settings of SilverEye, so that the swarm dynamics of that tool are represented in the console layers, Harris hawks optimization (HHO) (Heidari et al., 2019) and the peregrine falcon optimizer (PFO) (Wu and Liu, 2023). All three L-SHADE variants use the budget-aware schedule of Section 3.3, and HHO uses the horizon rescaling unless stated otherwise, because these are the configurations shipped in SilverHawk; the ablations in Section 6 switch both off. The portfolio selector of Section 3.4 is run with its defaults. In the travelling-salesman layer its results come from an earlier build of the selector and are reported for completeness only; on the shell the current build was used.

## 4.7 Statistical analysis

For each problem, the median and interquartile range of the final best value over the runs are reported. Solvers are compared across the problems of a set by their mean rank, computed on the per-problem medians, with the Friedman test (Derrac et al., 2011). Pairwise comparisons on a single problem use the two-sided Mann–Whitney U test at $\alpha = 0.05$. Where a solver is compared with all others, the p-values are adjusted with Holm's procedure (Holm, 1979). Pairwise results are summarized as wins, ties and losses (W/T/L) of one solver or configuration against another over the problems of a set: a win (W) or a loss (L) requires a significant difference in the stated direction, and a tie (T) means that the test finds none. This summary is used for MERLIN against the other solvers (Table 3, Holm-corrected), for the ablations (Section 6, uncorrected, one planned comparison per problem) and for the solvers of SilverHawk against the external tools (Section 5.6, uncorrected). To study how the ranking changes with the budget, the best-so-far value is also recorded after 25, 50, 75 and 100 % of the evaluations (console) or of the wall-clock time (shell).

On the shell, with five runs per solver, the smallest two-sided p-value that the Mann–Whitney test can produce is 0.008, so a correction over all pairs of sixteen solvers would make every difference non-significant by construction. Confirmatory testing is therefore restricted to a family of six comparisons and corrected only within it (Holm): MERLIN against CMA-ES, its own core; MERLIN against L-SHADE, the reference adaptive DE; MERLIN against each of the three external tools of the main campaign (Galapagos, Opossum-CMA-ES and Opossum-RBFOpt); and CMA-ES against Opossum-CMA-ES, which isolates the implementation of the same algorithm. The family reflects the questions of the paper but was not registered in advance: it was defined during the campaign, when the final values of CMA-ES, MERLIN and L-SHADE and the runs of the external tools were already known. Its confirmatory value is therefore limited, and for the comparisons with the external tools the evidence rests mainly on the complete separation of the samples, which the run-by-run panel of the shell figure (Section 5.5) shows. All other shell comparisons are reported as descriptive; this includes TPE and MADS, which were run after the family was defined.

All experiments were run on one workstation (Intel Core i7-4770K, 3.5 GHz, 32 GB RAM, Windows). The console experiments used .NET 8 and a single build of the console. The Grasshopper experiments used Rhinoceros 8 (SR35) and, for the shell, Karamba3D 3.1.60921.0; the travelling-salesman runs of SilverHawk were made between 23 March and 7 April 2026. The shell runs of SilverHawk were made on 14–15 October 2026 with the current build of the plug-in (v0.6.31), those of Galapagos, Opossum-CMA-ES and Opossum-RBFOpt on 9–11 May 2026 with the Galapagos version shipped with Rhinoceros 8 and Opossum 3.0.1, and those of Opossum TPE and MADS on 15 October 2026 with Opossum 3.2.1, all on the same definition and machine.

**Table 1.** Problem classes, budgets and number of runs per solver.

| Layer | Problems | D | Variables | Budget | Runs/solver | Environment |
|---|---|---|---|---|---|---|
| F1 synthetic, original | Sphere, Rosenbrock, Rastrigin, Ackley, Griewank, Schwefel 2.26, Schwefel 1.2, Step | 10, 30 | continuous | 1,200 FE (hard cap) | 30 | console |
| F2 synthetic, shifted–rotated | as F1 without Schwefel 2.26 | 10, 30 | continuous | 1,200 FE (hard cap) | 30 | console |
| TSP | 10, 15, 20 cities, branch encoding (integer sliders) | 10–20 | integer → tour (many-to-one) | 1,200 FE | 10 (Tunny 3) | Grasshopper |
| Building performance | Q_tot, K_h, DF (one objective at a time) | 22 | 18 continuous + 4 binary | 1,200 FE (hard cap) | 30 | console (surrogate) |
| Shell (Karamba3D) | maximum displacement under self-weight | 85 | continuous | 15 min wall-clock | 5 | Grasshopper |

**Table 2.** Solvers compared. Reference algorithms use their published settings; external tools use the defaults of the installed version.

| Solver | Family | Role | Layers |
|---|---|---|---|
| MERLIN | CMA-ES + DE injection (+ Solis–Wets) | proposed (Sec. 3.1) | all |
| RAPTOR | PSO + success history + Lévy + DE injection (+ Solis–Wets) | proposed (Sec. 3.2) | all |
| L-SHADE-SW | L-SHADE + Solis–Wets, budget-aware LPSR | proposed (Sec. 3.3) | all |
| L-SHADE | adaptive DE, budget-aware LPSR | reference + schedule | all |
| L-SHADE-RSP | adaptive DE with rank-based selective pressure, budget-aware LPSR | reference variant + schedule | all |
| CMA-ES | evolution strategy | reference | all |
| jEDE | self-adaptive DE, strategy ensemble (as in Optimus) | reference | all |
| PSO-IW | particle swarm, SilverEye settings | reference | all |
| HHO | swarm (Harris hawks) | reference | all |
| PFO | swarm (peregrine falcon) | reference | all |
| Portfolio selector | algorithm selection | reference row | all |
| Galapagos GA | genetic algorithm | external tool | TSP, shell |
| Opossum RBFOpt / CMA-ES | surrogate model / evolution strategy | external tool | TSP, shell |
| Opossum TPE / MADS (3.2.1) | tree-structured Parzen estimator / mesh adaptive direct search | external tool | shell |
| SilverEye | particle swarm | external tool | TSP |
| Tunny TPE | Bayesian (tree-structured Parzen) | external tool | TSP |

## 5. Results

### 5.1 Synthetic functions: the effect of shifting and rotating

Figure 4 summarizes the synthetic results as mean ranks of the per-function medians. On the original, centred functions the picture is mixed. At D = 10 the portfolio selector (2.5) and MERLIN (2.6) share the lead. At D = 30 Harris hawks optimization is first with a mean rank of 1.6, ahead of the selector (2.5) and MERLIN (3.0). Read in isolation, this ranking would recommend HHO for 30-variable problems.

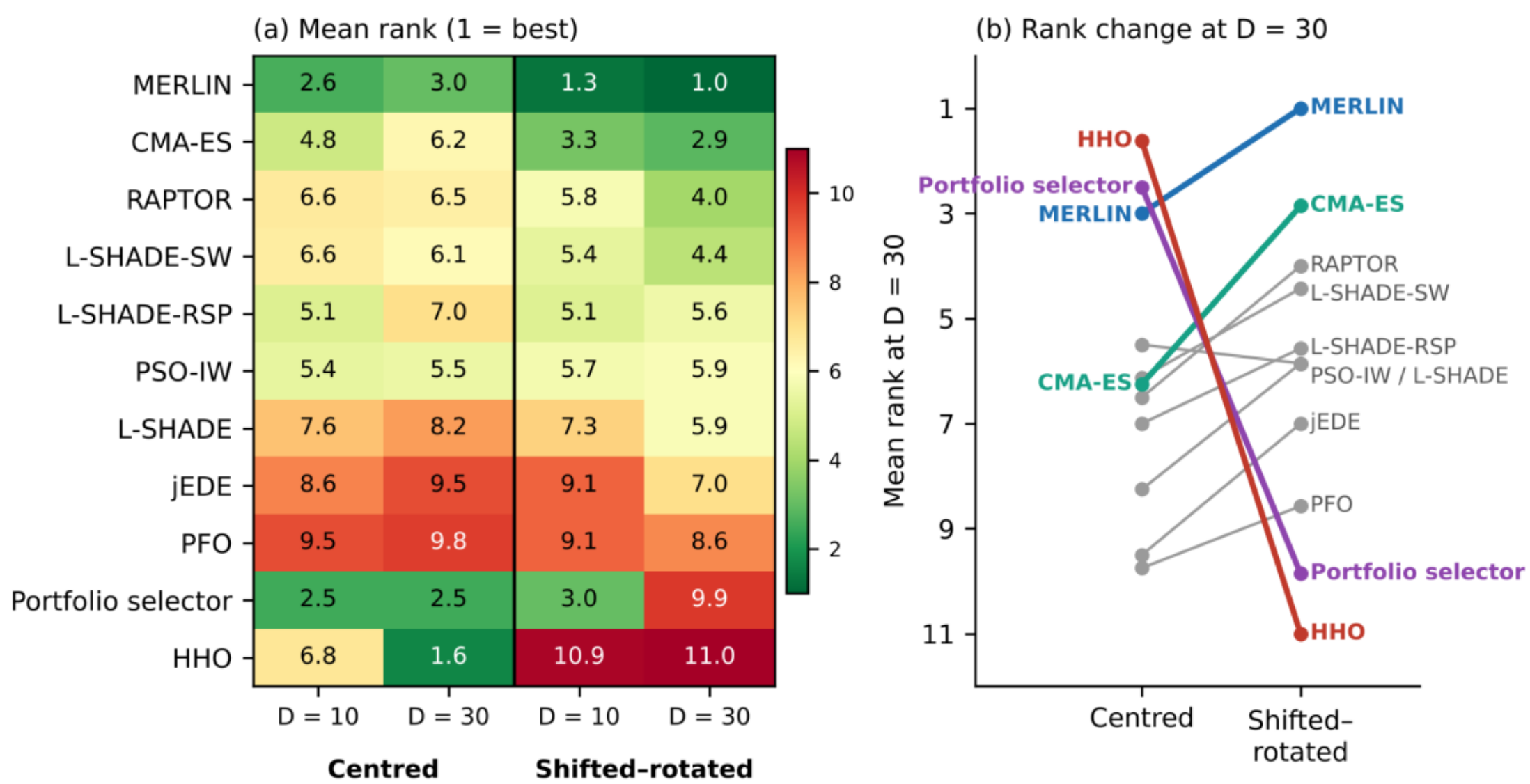

**Figure 4.** Effect of the centre bias. (a) Mean rank of the per-function medians, 1,200 evaluations, 30 runs, on the centred (F1) and shifted–rotated (F2) synthetic functions at D = 10 and D = 30. (b) Change of the mean rank at D = 30 between the two sets.

Shifting and rotating the same functions changes that recommendation completely. On the F2 set HHO falls to last place at both dimensions (10.9 and 11.0 out of 11), and the selector, which relies on HHO as one of its arms, drops to 9.9 at D = 30. The size of the reversal is visible on single functions. On centred Rastrigin at D = 30 the median of HHO is 10.5, one of the best values of the roster; on the shifted and rotated version it is 339, the worst. HHO had found the centre of the box, not the optimum of the function. This is the structural bias described by Kůdela (2022), here measured on an algorithm that is widely used in building-optimization studies, and it is the reason why all further synthetic comparisons in this paper use the F2 set.

On F2 MERLIN is first at both dimensions (mean rank 1.29 at D = 10 and 1.00 at D = 30), followed by CMA-ES (3.29 and 2.86). At D = 30 MERLIN has the best median on all seven functions. Against each of the other ten solvers it wins on six or seven of the seven functions after Holm correction and loses on none (Table 3). Against CMA-ES, which shares its sampling core, the record is 4/3/0 at D = 10 and 6/1/0 at D = 30; Section 6 attributes this margin to the differential-evolution injection. Among the remaining solvers, RAPTOR (4.0) and L-SHADE-SW (4.4) are third and fourth at D = 30, ahead of the reference adaptive DE variants.

**Table 3.** Shifted–rotated functions, D = 30, 1,200 evaluations: median of the final best value over 30 runs. Bold: best median. W/T/L: wins, ties and losses of MERLIN against the solver over the seven functions (Mann–Whitney, Holm-corrected, α = 0.05). The D = 10 table and the interquartile ranges are in the supplementary material.

| Solver | Ackley | Griewank | Rastrigin | Rosenbrock | Schwefel 1.2 | Sphere | Step | W/T/L |
|---|---|---|---|---|---|---|---|---|
| MERLIN | **3.84** | **1.70** | **170** | **246** | **1.39e4** | **0.213** | **77.5** | – |
| CMA-ES | 4.88 | 2.77 | 224 | 263 | 2.21e4 | 0.466 | 158 | 6/1/0 |
| RAPTOR | 15.7 | 6.11 | 218 | 1.74e3 | 2.02e4 | 1.25 | 732 | 7/0/0 |
| L-SHADE-SW | 12.1 | 13.7 | 230 | 3.12e3 | 1.84e4 | 3.95 | 1.14e3 | 7/0/0 |
| L-SHADE-RSP | 11.9 | 19.8 | 269 | 3.90e3 | 1.62e4 | 6.30 | 2.21e3 | 6/1/0 |
| L-SHADE | 11.5 | 27.1 | 262 | 5.63e3 | 1.74e4 | 6.23 | 2.15e3 | 7/0/0 |
| PSO-IW | 12.1 | 53.9 | 242 | 4.56e3 | 1.73e4 | 14.8 | 2.08e3 | 7/0/0 |
| jEDE | 12.3 | 22.1 | 282 | 5.66e3 | 3.21e4 | 5.92 | 1.93e3 | 7/0/0 |
| PFO | 18.3 | 68.3 | 257 | 3.27e4 | 3.14e4 | 20.5 | 8.43e3 | 7/0/0 |
| Portfolio selector | 17.5 | 93.6 | 286 | 3.29e4 | 4.41e4 | 32.1 | 9.51e3 | 7/0/0 |
| HHO | 20.0 | 299 | 339 | 2.27e5 | 1.36e5 | 88.1 | 2.76e4 | 7/0/0 |

The values at D = 30 are far from the optima. With 1,200 evaluations, about 40 per variable, no solver solves these functions, and the comparison measures how much progress each method makes in a budget that a designer would consider generous for a simulation-based problem. At D = 10 the values are closer to the optima (MERLIN reaches 0.03 on Ackley and 0.11 on Griewank), and CMA-ES and the selector also solve Sphere and Step.

## 5.2 The budget horizon

Final values hide how a ranking develops during a run. Figure 5 shows the mean rank at 25, 50, 75 and 100 % of the budget. On the shifted–rotated functions at D = 30 (Fig. 5a) MERLIN leads from the first quarter of the run and never loses the lead. The selector is third after a quarter of the budget and ninth or tenth from the midpoint onwards. RAPTOR and L-SHADE-SW improve their position steadily as the budget is spent.

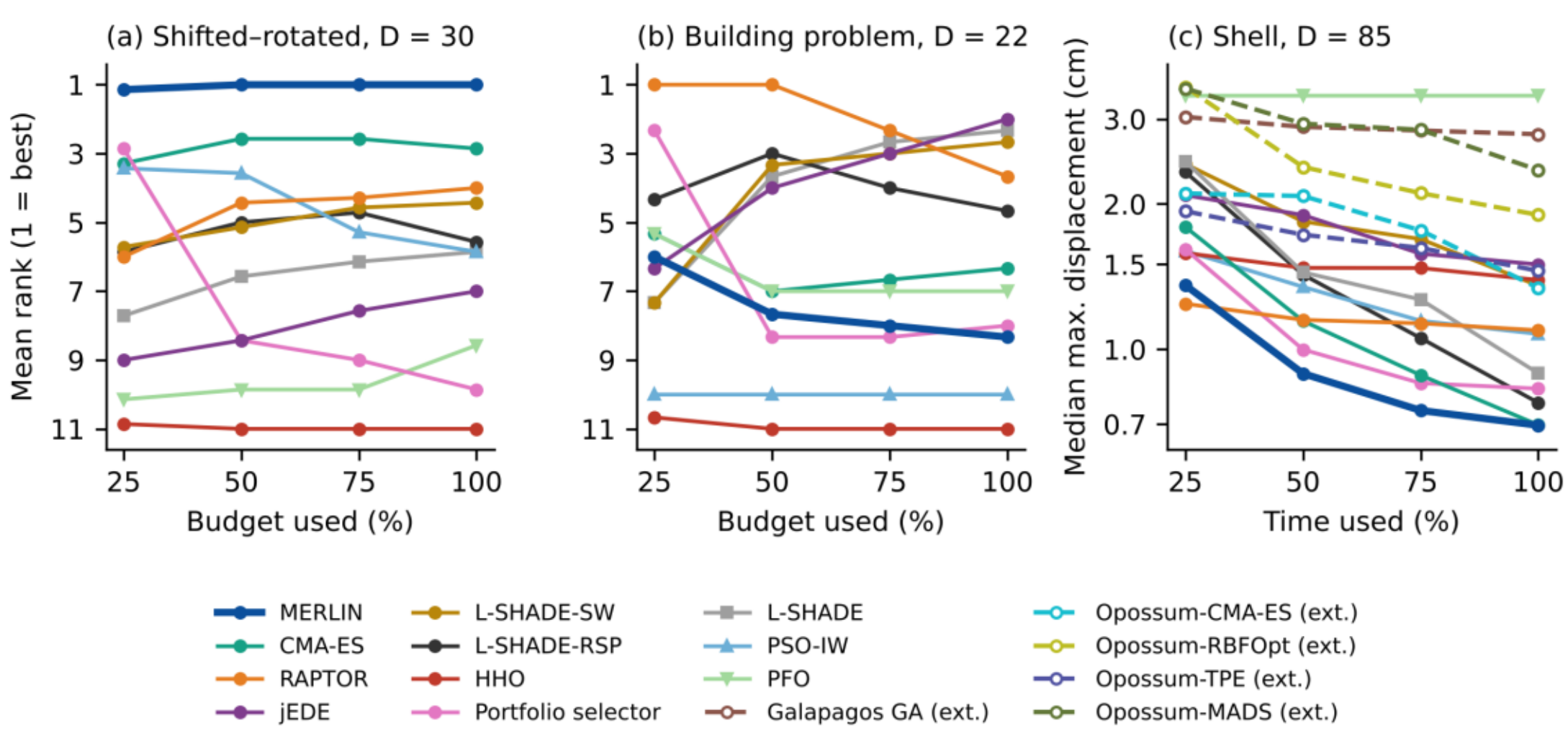

**Figure 5.** The budget horizon. (a, b) Mean rank after 25, 50, 75 and 100 % of the 1,200 evaluations on the shifted–rotated functions at D = 30 and on the building problem (three objectives). (c) Shell: median best-so-far displacement after 25, 50, 75 and 100 % of the 15 minutes, 5 runs per solver; for Galapagos and the Opossum 3.0.1 solvers, whose logs record evaluations, time is approximated by the fraction of evaluations completed; the logs of Opossum-TPE and Opossum-MADS record time.

The building problem (Fig. 5b, Section 5.4) shows a different pattern, and the shell (Fig. 5c, Section 5.5) a third one. Taken together, the three panels show that the ranking of solvers depends on where the run is stopped as much as on the problem. A designer who stops a run early to inspect intermediate designs effectively uses a different optimizer from one who lets it finish.

## 5.3 Travelling-salesman instances

Table 4 and Figure 6 report the travelling-salesman results (Fig. 6a–c, gaps; Fig. 6d–f, best tours on the 20-city instance). The 10-city instance is too easy to separate the solvers: five of them, two external, have a median equal to the optimum, and only HHO is significantly worse than the best after Holm correction. Differences appear at 15 and 20 cities. CMA-ES has the best median on both (44.78 and 54.29), MERLIN is second among the solvers of SilverHawk (50.63 and 55.63), and the two are not distinguishable after Holm correction ($p = 0.10$ and 0.62). With ten runs per solver the tests have limited power: the two Opossum solvers are not separable from CMA-ES either, although their medians are 30–35 % above the optimum on the 20-city instance. MERLIN also found the best tour of the whole campaign on the 20-city instance (49.32, 1.8 % above the optimum of 48.46), 1.8 % shorter than the best tour of any external tool (50.24, Opossum-RBFOpt; Fig. 6d–e). The two tours share 17 of 20 edges and differ only in the connections around the cities F, M, P, Q, J and A, whereas the best Galapagos tour (58.33, Fig. 6f) is 18 % longer and contains several long chords across the instance. Opossum-RBFOpt is the strongest external tool, with no significant difference from CMA-ES on any instance, but its median gap grows from 0 % to 4 % to 30 % as the instance grows from 10 to 20 cities. Its runs also take between 3 and 25 minutes, against about a minute for the population-based solvers, because the model fitting dominates the cost.

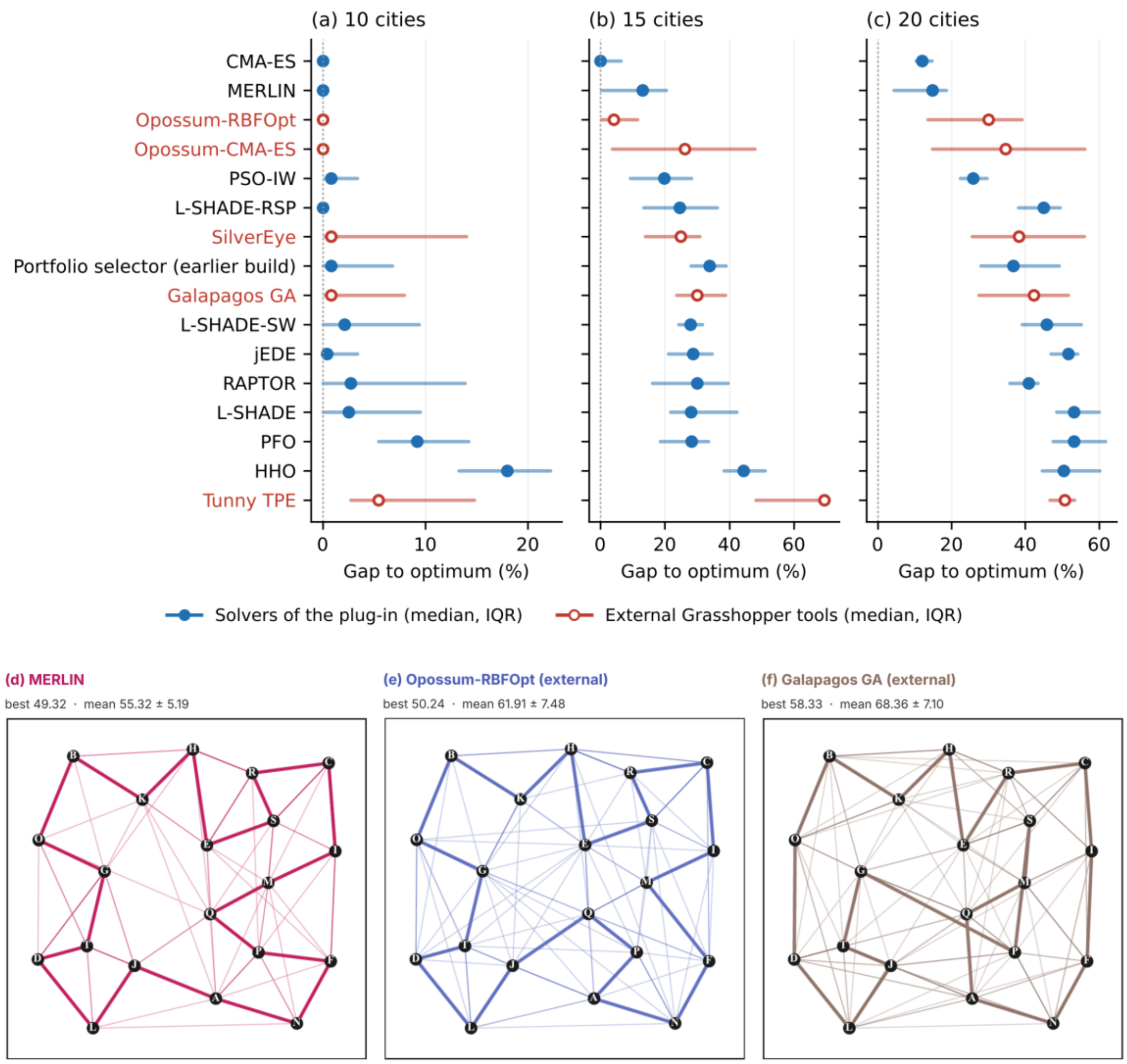

**Figure 6.** Travelling-salesman instances, 1,200 evaluations. (a–c) Median and interquartile range of the gap to the optimum (computed exactly with the Held–Karp algorithm), 10 runs per solver (Tunny 3); external tools in red. (d–f) 20 cities (optimum 48.46): best tour of the ten runs (bold) of MERLIN, of the best external tool (Opossum-RBFOpt) and of the standard Grasshopper tool (Galapagos GA); thin lines: edges of the routes sampled along the ten runs, with the same colour code. Tour lengths 49.32, 50.24 and 58.33; mean ± SD over the ten runs below each title.

The differential-evolution family and RAPTOR, which do well on the synthetic functions, fall to the lower half of the ranking on this problem: on the 20-city instance L-SHADE, jEDE and L-SHADE-SW have median gaps between 46 and 53 %. Their search relies on difference vectors between candidates, and on the plateaus of the branch encoding most differences either move no city or move many at once. CMA-ES and MERLIN adapt their sampling distribution to the plateau structure instead. HHO is last or second to last on all three instances. A control experiment with a bijective random-key encoding on the 10- and 15-city instances (Supplementary S4) leaves CMA-ES and MERLIN first and second on the 15-city instance, followed by Opossum-RBFOpt, so this ranking does not depend on the many-to-one branch encoding.

**Table 4.** Travelling-salesman instances, 1,200 evaluations: median (interquartile range) of the tour length and, for 20 cities, median gap to the optimum (exact, Held–Karp). Bold: best median. † external tool; Tunny 3 runs, all others 10. ns: not significantly different from the best median (Mann–Whitney on values rounded to 0.01, Holm, α = 0.05); on the 10-city instance all solvers except HHO are ns. Full table in the supplementary material.

| Solver | 10 cities (opt. 39.06) | 15 cities (opt. 44.78) | 20 cities (opt. 48.46) |
|---|---|---|---|
| CMA-ES | **39.06** (39.06–39.06) | **44.78** (44.78–47.67) | **54.29** (53.50–55.61), gap 12 % |
| MERLIN | **39.06** (39.06–39.06) | 50.63 (44.98–53.97) ns | 55.63 (50.59–57.45) ns, 15 % |
| Opossum-RBFOpt † | **39.06** (39.06–39.06) | 46.63 (44.98–49.94) ns | 63.00 (55.01–67.38) ns, 30 % |
| Opossum-CMA-ES † | **39.06** (39.06–39.06) | 56.50 (46.42–66.19) ns | 65.22 (55.61–75.63) ns, 35 % |
| PSO-IW | 39.39 (39.14–40.39) | 53.62 (48.98–57.42) | 61.01 (59.28–62.80), 26 % |
| L-SHADE-RSP | **39.06** (39.06–39.06) | 55.78 (50.83–60.99) | 70.27 (66.90–72.40), 45 % |
| SilverEye † | 39.39 (39.14–44.54) | 55.96 (51.03–58.61) | 66.99 (60.87–75.58), 38 % |
| Galapagos GA † | 39.39 (39.14–42.17) | 58.21 (55.36–62.14) | 68.95 (61.75–73.50), 42 % |
| RAPTOR | 40.12 (39.06–44.48) | 58.23 (51.99–62.48) | 68.30 (65.77–69.47), 41 % |
| L-SHADE-SW | 39.89 (39.06–42.73) | 57.29 (55.66–58.92) | 70.62 (67.40–75.16), 46 % |
| jEDE | 39.23 (39.06–40.39) | 57.63 (54.24–60.26) | 73.46 (71.19–74.70), 52 % |
| L-SHADE | 40.05 (39.06–42.77) | 57.35 (54.52–63.69) | 74.20 (71.90–77.56), 53 % |
| PFO | 42.66 (41.18–44.62) | 57.42 (53.07–59.77) | 74.26 (71.45–78.33), 53 % |
| HHO | 46.09 (44.24–47.73) | 64.67 (61.93–67.62) | 72.89 (70.05–77.58), 50 % |
| Tunny TPE † | 41.18 (40.12–44.84) | 75.88 (66.39–76.33) | 73.00 (71.04–74.28), 51 % |
| Portfolio selector (earlier build) | 39.39 (39.06–41.72) | 59.95 (57.40–62.19) | 66.27 (62.00–72.27), 37 % |

## 5.4 Building-performance problem

On the building problem the ranking changes again (Table 5). The differential-evolution solvers lead: jEDE has the best median on Q_tot and DF, and L-SHADE-SW on K_h. The best group is narrow for DF, where only jEDE is significantly better than all others, and broad for K_h, where eight solvers, MERLIN included, cannot be separated from the best after Holm correction. For Q_tot the group contains jEDE, RAPTOR, L-SHADE and L-SHADE-SW. MERLIN, first on every rotated synthetic function, is eighth on Q_tot and K_h and ninth on DF by median; HHO and PSO-IW are last on all three objectives.

Two features of the problem plausibly explain the change. Four of its 22 variables are binary and enter the model through rounding, which creates discontinuities that a smooth covariance model cannot represent, whereas differential evolution moves along the coordinate axes through binomial crossover. Moreover, the three objectives depend strongly on few variables (insulation, air-change rate and sill height in the case of energy), so the problem is closer to separable than the rotated functions, a structure on which DE crossover is known to be efficient.

The budget horizon is pronounced here (Fig. 5b). RAPTOR is first after 25 % and 50 % of the budget, and the DE solvers overtake it only in the second half. MERLIN starts mid-table and loses positions as the budget is spent. A designer who can afford only a few hundred evaluations of this problem would be better served by RAPTOR, one who can afford a thousand by jEDE or L-SHADE-SW.

**Table 5.** Building-performance problem (D = 22), 1,200 evaluations, 30 runs: median (interquartile range) of the final best value for each objective, minimized (Q_tot in kWh $m^{-2}$ $a^{-1}$, K_h in h) or maximized (DF). Bold: best median; ns: not significantly different from it (Mann–Whitney, Holm, $\alpha = 0.05$).

| Solver | Q_tot | K_h | DF |
|---|---|---|---|
| jEDE | **12.03** (11.93–12.08) | 18.16 (17.37–18.71) ns | **0.372** (0.370–0.374) |
| RAPTOR | 12.04 (11.88–12.33) ns | 18.75 (17.22–20.51) ns | 0.367 (0.362–0.371) |
| L-SHADE | 12.06 (11.93–12.15) ns | 17.75 (17.06–19.29) ns | 0.369 (0.366–0.370) |
| L-SHADE-SW | 12.11 (11.99–12.27) ns | **17.30** (16.57–18.96) | 0.368 (0.365–0.371) |
| CMA-ES | 12.11 (12.00–12.26) | 18.91 (18.09–20.36) ns | 0.359 (0.358–0.361) |
| L-SHADE-RSP | 12.15 (12.00–12.34) | 17.76 (16.58–18.56) ns | 0.367 (0.361–0.371) |
| PFO | 12.19 (12.05–12.32) | 19.88 (18.42–22.81) | 0.366 (0.362–0.369) |
| MERLIN | 12.23 (12.09–12.32) | 19.89 (18.31–21.31) ns | 0.357 (0.354–0.360) |
| Portfolio selector | 12.29 (12.07–12.52) | 20.75 (17.76–33.23) ns | 0.367 (0.361–0.371) |
| PSO-IW | 12.71 (12.51–12.88) | 29.55 (25.61–35.18) | 0.357 (0.352–0.361) |
| HHO | 13.30 (12.86–13.82) | 33.65 (31.31–37.92) | 0.339 (0.314–0.350) |

## 5.5 Free-form shell

Starting from the flat surface, whose maximum displacement under self-weight is 85.4 cm, every solver reduces the displacement by more than an order of magnitude within 15 minutes (Table 6, Fig. 7). The two best solvers end at a median of 0.70 cm, a reduction of more than 99 %, and nine of the eleven solvers of SilverHawk finish below 1.5 cm. Figure 7a shows what this means for the surface. On the colour scale of the flat start (0–85.4 cm) the best run of MERLIN, whose maximum displacement is 0.50 cm, 170 times lower than the starting value, is uniformly at the bottom of the scale, and its displacement field becomes visible only on its own scale (0–0.50 cm); the deformation is magnified 25 times in all three views. Figure 7b gives the final value of every run.

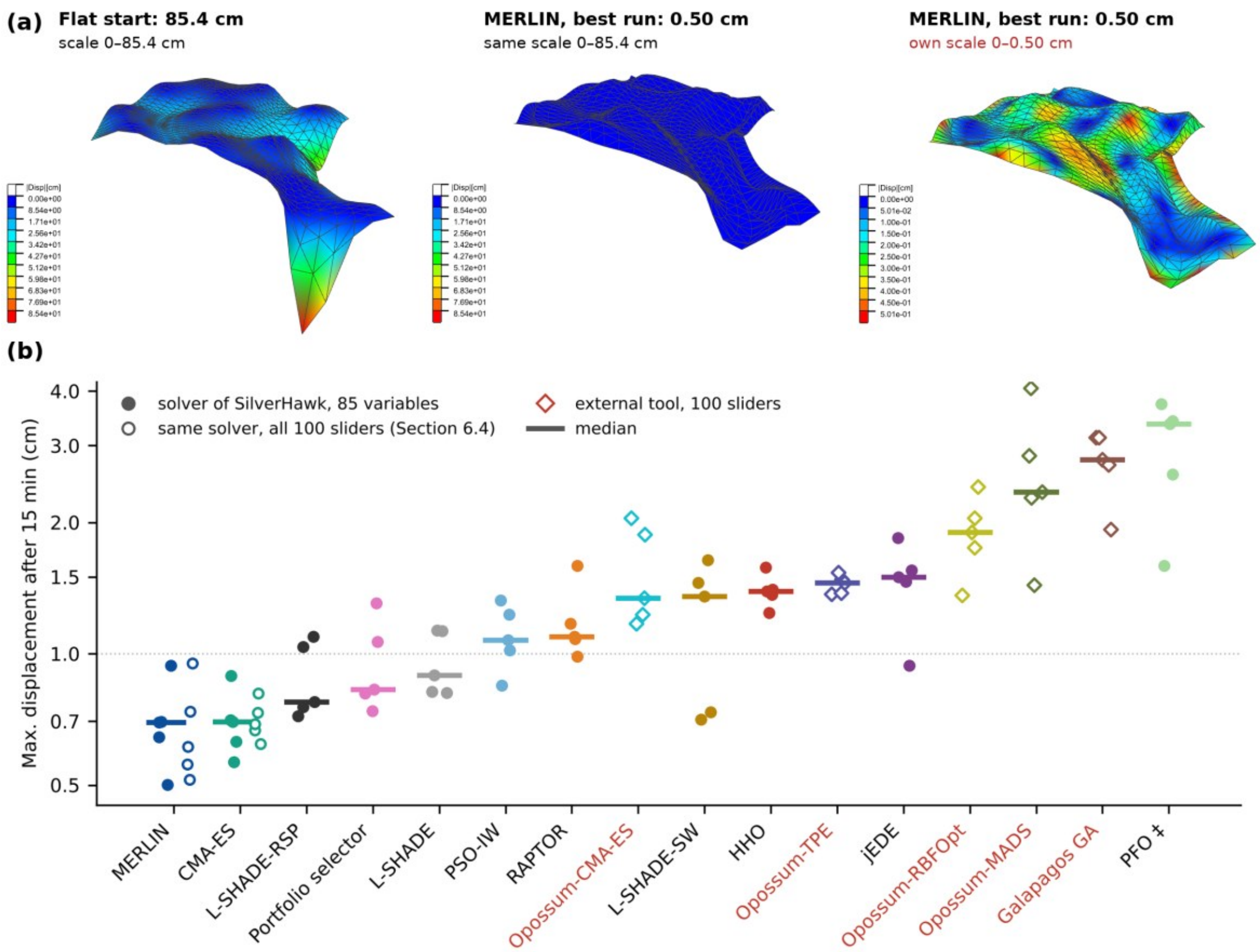

**Figure 7.** Free-form shell. (a) Displacement magnitude |Disp| (Karamba3D, cm) of the flat starting surface (maximum 85.4 cm) and of the best run of MERLIN (maximum 0.50 cm) on the same colour scale, 0–85.4 cm, and of the same run on its own scale, 0–0.50 cm; the deformation is magnified 25 times in all three views. (b) Final maximum displacement of every run after 15 minutes (flat starting surface 85.4 cm). Filled circles: solvers of SilverHawk with the 85 effective variables; open circles: CMA-ES and MERLIN with all 100 sliders (Section 6.4); open diamonds: external tools; bars: medians. ‡ PFO stops by stagnation and is not time-fair.

MERLIN and CMA-ES share the first position: their medians are 0.696 and 0.697 cm and the samples cannot be separated ($p = 1.0$). L-SHADE-RSP (0.775 cm) and the portfolio selector (0.829 cm) follow, then L-SHADE (0.892 cm), which is the closest to MERLIN among the comparisons of the corrected family without a significant difference ($p = 0.056$, $p_{Holm} = 0.11$). The five external tools end behind the leaders: Opossum-CMA-ES at 1.34 cm, Opossum-TPE at 1.45 cm, Opossum-RBFOpt at 1.90 cm with the 418 evaluations that fit into its time budget, Opossum-MADS at 2.35 cm and Galapagos at 2.79 cm. The three comparisons of MERLIN with an external tool and that of CMA-ES with Opossum-CMA-ES, all in the corrected family, are significant after Holm correction ($p_{Holm} = 0.048$ for each), and in every case the five runs of the SilverHawk solver are all better than the five runs of the external tool. The same complete separation holds, outside the corrected family, between MERLIN or CMA-ES and TPE or MADS ($p = 0.008$). L-SHADE-SW is bimodal on this problem: two runs end near the leaders (0.71 and 0.74 cm) and three between 1.35 and 1.64 cm.

The results must be read with the conditions of Section 3.5 in mind: in this layer MERLIN ran without its Solis–Wets step and the L-SHADE variants with a constant population, so the tie at the top is between CMA-ES with DE injection and CMA-ES alone. The injection, which separates MERLIN from CMA-ES on the rotated functions, does not separate them here at the end of the run. It does earlier: the anytime medians below show MERLIN ahead during the whole run, and at a fixed count of 1,943 evaluations, the smallest number reached by any time-fair run, MERLIN is at 0.726 cm and CMA-ES at 0.800 cm. CMA-ES is still improving when the time runs out (0.83 cm at 2,000 evaluations, 0.70 cm at the end), so a longer budget may widen or close this gap; with five runs neither is claimed.

The time axis (Fig. 5c) repeats in wall-clock form the pattern of the building problem. After a quarter of the time RAPTOR has the lowest median (1.24 cm), followed by MERLIN (1.36 cm), while CMA-ES and L-SHADE-RSP are still at 1.79 and 2.33 cm. MERLIN takes the lead at the midpoint (0.89 cm) and keeps it to the end; CMA-ES closes the gap only in the last quarter. HHO and PSO-IW, competitive in the first quarter, improve little afterwards and end at 1.39 and 1.08 cm. The median time to reach 1 cm is 373 s for MERLIN and 548 s for CMA-ES; HHO, jEDE and PFO do not reach it in most runs. The external tools stay above the leading solvers of SilverHawk throughout: Opossum-CMA-ES improves mainly in the last quarter (2.10 to 1.34 cm), Opossum-RBFOpt from 3.50 to 1.90 cm, Opossum-TPE from 1.93 to 1.45 cm, Opossum-MADS from 3.47 to 2.35 cm, and Galapagos hardly moves after the first quarter (3.03 to 2.79 cm).

The time limit decides the comparison with Opossum-RBFOpt, which spends about 1.75 s of model fitting per evaluation on top of the 0.4 s of the analysis. At an equal number of evaluations the picture is different. After 418 evaluations the median of MERLIN is 1.71 cm and that of RAPTOR 1.28 cm, neither significantly different from RBFOpt (1.90 cm; $p = 0.31$ and 0.22), while CMA-ES (2.47 cm) and L-SHADE-RSP (2.62 cm) are significantly worse than RBFOpt at that point ($p = 0.032$ and 0.016, uncorrected). The surrogate-based solver therefore uses evaluations at least as efficiently as the best population-based solvers in this range; it loses on the shell because, with a fast analysis, its own overhead costs more than the evaluations it saves. Section 7.3 returns to this crossover.

Of the two solvers added in Opossum 3.2, TPE comes closer to the leaders, but it is held back by its own overhead: it completes only 1,046–1,100 evaluations in 15 minutes, about half as many as the solvers of SilverHawk, because its model is refitted to a growing set of trials. Per evaluation it is as efficient as the other model-based tool: after 418 evaluations its median (1.91 cm) equals that of Opossum-RBFOpt (1.90 cm; $p = 0.84$) and is below that of CMA-ES (2.47 cm). At the end of the run it cannot be separated from Opossum-CMA-ES ($p = 0.69$) or Opossum-RBFOpt ($p = 0.15$). MADS keeps a constant cost per evaluation but is still descending at 15 minutes, and it improves further after 900 s in every run; it is worse than Opossum-CMA-ES ($p = 0.032$) and cannot be separated from Opossum-RBFOpt or Galapagos.

Cichocka et al. (2017) solved the same structure with SilverEye, Galapagos and Octopus under 15-minute and 8-hour limits. In 15 minutes the best average they report is 1.79 cm (SilverEye with its best velocity setting), against 4.25 cm for Octopus and 4.74–5.36 cm for Galapagos; after 8 hours SilverEye reached 0.6 cm, Octopus 1.6 cm and Galapagos 2.8 cm. In the 15-minute runs of this study the medians of the two leading solvers are 0.70 cm. The comparison is indicative only: the shell thickness is the same (20 cm), but their variables ranged over [−10, 10] m instead of [−5, 5] m, and the finite-element model, the Karamba version and the hardware differ. The solvers of this study are therefore not ranked against those results, but the order of magnitude is clear: values that took hours with the tools available in 2017 are approached here in a quarter of an hour.

**Table 6.** Free-form shell (D = 85), 15 minutes per run, 5 runs per solver: maximum displacement (cm); flat starting surface 85.4 cm. Median (interquartile range) and mean of the final value; median best value after 1,943 evaluations, the smallest count reached by a time-fair run; median number of evaluations per run; two-sided Mann–Whitney p-value against MERLIN. Bold: best median. * in the family of six corrected comparisons (Section 4.7), significant after Holm correction; other p-values are descriptive. † external tool (100 sliders, 85 effective). ‡ stops by stagnation after 1–3 minutes, not time-fair. Opossum-TPE and Opossum-MADS: Opossum 3.2.1, 15 October 2026, best value within 900 s of the log (Section 4.5).

| Solver | Median (IQR) | Mean | At 1,943 FE | FE | p vs MERLIN |
|---|---|---|---|---|---|
| MERLIN | **0.696** (0.644–0.698) | 0.696 | **0.726** | 2,405 | – |
| CMA-ES | 0.697 (0.628–0.704) | 0.697 | 0.800 | 2,296 | 1.000 (family) |
| L-SHADE-RSP | 0.775 (0.756–1.037) | 0.876 | 0.920 | 2,300 | 0.056 |
| Portfolio selector | 0.829 (0.812–1.066) | 0.950 | 0.830 | 2,219 | 0.056 |
| L-SHADE | 0.892 (0.819–1.127) | 0.957 | 1.072 | 2,258 | 0.056 (family) |
| PSO-IW | 1.076 (1.019–1.231) | 1.099 | 1.099 | 2,124 | 0.016 |
| RAPTOR | 1.095 (1.082–1.174) | 1.186 | 1.096 | 2,116 | 0.008 |
| Opossum-CMA-ES † | 1.340 (1.229–1.876) | 1.533 | – | 2,370 | 0.008* |
| L-SHADE-SW | 1.355 (0.736–1.455) | 1.178 | 1.449 | 2,181 | 0.032 |
| HHO | 1.392 (1.365–1.403) | 1.396 | 1.406 | 2,276 | 0.008 |
| Opossum-TPE † | 1.454 (1.377–1.481) | 1.444 | – | 1,092 | 0.008 |
| jEDE | 1.499 (1.464–1.556) | 1.461 | 1.556 | 2,235 | 0.008 |
| Opossum-RBFOpt † | 1.899 (1.754–2.045) | 1.894 | – | 418 | 0.008* |
| Opossum-MADS † | 2.352 (2.279–2.846) | 2.597 | – | 1,539 | 0.008 |
| Galapagos GA † | 2.790 (2.716–3.135) | 2.741 | – | 2,414 | 0.012* |
| PFO ‡ | 3.363 (2.577–3.412) | 2.935 | – | – | 0.008 |

## 5.6 The solvers of SilverHawk against the external tools

Sections 5.3 and 5.5 compare all solvers against the best one. A designer choosing a tool asks a different question: is a given solver better than the tool already in use? Figure 8 answers it for every pair of a solver of SilverHawk and an external tool on the four problems run inside Grasshopper, with a two-sided Mann–Whitney test per pair ($\alpha = 0.05$, uncorrected; the Holm-corrected comparisons against the best solver are those of Tables 4 and 6).

Two solvers are never significantly worse than any external tool. CMA-ES wins 15 of 20 comparisons and MERLIN 13, and neither loses one. Both are better than SilverEye and Galapagos on every instance and better than Tunny on the two larger travelling-salesman instances. On the shell they are better than all five external tools, as are L-SHADE-RSP, L-SHADE and the portfolio selector. The medians of these five solvers (0.70–0.89 cm) are 33–48 % below that of the best external tool, Opossum-CMA-ES (1.34 cm), and at most a third of that of Galapagos (2.79 cm). Galapagos is beaten on the shell by every solver of SilverHawk except PFO, and the only shell losses in the figure are those of PFO against Opossum-CMA-ES and Opossum-TPE. With five runs per solver, the shell wins correspond to a complete or near-complete separation of the samples rather than a difference in the tails; for the four comparisons with external tools in the corrected family the separation is complete and survives Holm correction (Section 5.5).

The one external tool that competes is Opossum-RBFOpt on the travelling-salesman instances. It is better than most solvers of SilverHawk at 10 and 15 cities and than six of them at 20 cities, and only CMA-ES beats it, at 20 cities; MERLIN ties with it on all three instances ($p$ = 0.054 at 20 cities). This strength has two limits. Its advantage shrinks as the instance grows (median gap 0 %, 4 % and 30 %, Section 5.3), and each of its runs takes between 3 and 25 minutes against about one minute for the population-based solvers. On the shell, where the time limit restricts it to 418 evaluations, it is beaten by eight solvers of SilverHawk, although at an equal number of evaluations it is not worse than any of them (Section 5.5).

Finally, the comparison of CMA-ES with Opossum-CMA-ES isolates the effect of implementation and integration, since both run the same algorithm family with default settings. The solver of SilverHawk is better on the 15-city instance and on the shell (0.70 against 1.34 cm, complete separation, $p_{Holm}$ = 0.048) and never significantly worse, so the advantage of the plug-in does not come only from new algorithms.

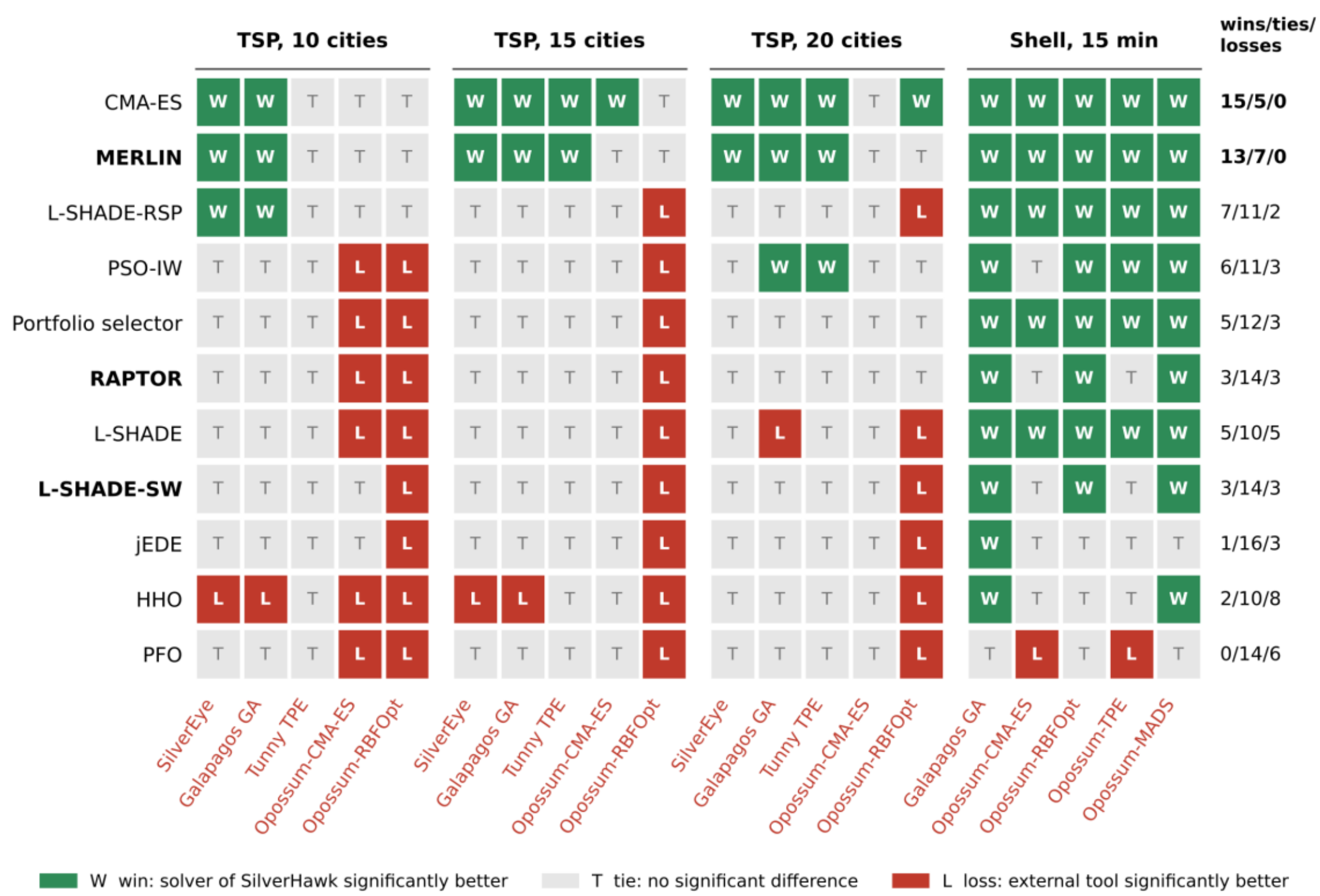

**Figure 8.** Every solver of SilverHawk (rows) against every external Grasshopper tool (columns) on the problems run inside Grasshopper: W, the solver of SilverHawk is better; L, the external tool is better; T, no significant difference (two-sided Mann–Whitney, $\alpha$ = 0.05, uncorrected). Travelling-salesman instances: 10 runs per solver, Tunny 3. Shell: 5 runs per

solver and per external tool. Opossum-TPE and Opossum-MADS: Opossum 3.2.1. Right: totals over the 20 comparisons.

## 6. Ablation: which components earn their evaluations

Each component of the proposed algorithms was removed or changed in turn, keeping everything else fixed, and the variant was run under the same hard cap of 1,200 evaluations on the seven shifted–rotated functions at $D = 10$ and $D = 30$ and on the three objectives of the building problem. Because the cap counts every evaluation, a component that consumes evaluations without improving the search shows up as a loss rather than as free extra effort. For each problem the full algorithm and the variant are compared with a two-sided Mann–Whitney test at $\alpha = 0.05$; each comparison is planned in advance and is reported uncorrected as a win, tie or loss of the full algorithm (W/T/L). Figure 9 shows, for each problem, the ratio between the median of the variant and that of the full algorithm, so that the size of the effects can be read next to their significance.

### 6.1 MERLIN

**Differential-evolution injection.** Removing the injected offspring turns MERLIN into CMA-ES with the same initial step size. MERLIN wins on four of seven functions at $D = 10$ and six of seven at $D = 30$, with no loss (Fig. 9a), and the medians of the variant are up to an order of magnitude worse on Rosenbrock and Step. Changing the initial step size from 0.1 to the usual 0.3 makes no difference on any problem (0/7/0, 0/7/0, 0/3/0). The advantage of MERLIN over CMA-ES on rotated landscapes therefore comes from the injection and not from the smaller step size. On the building problem the injection brings nothing: two ties and one loss, consistent with the poor position of both MERLIN and CMA-ES on that problem (Section 5.4).

**Injection fraction.** The best fraction depends on the dimension. At $D = 10$ the default of 20 % beats 10 %, 30 % and 50 % (3/4/0, 2/5/0, 6/1/0). At $D = 30$, 10 % is better than 20 % (1/1/5), while larger fractions are clearly worse (5/2/0 and 6/0/1 for the full algorithm against 30 % and 50 %). A plausible reading is that the injected vectors displace Gaussian samples from the covariance update: with $\lambda = 20$ and $D = 30$ the distribution already has few samples per estimated parameter, and every injected vector reduces that number further.

**Solis–Wets refinement.** Removing the Solis–Wets step does not hurt MERLIN anywhere and improves it at $D = 30$ (1/1/5 for the full algorithm), while $D = 10$ and the building problem are close to neutral (1/4/2 and 0/3/0). The step costs about five generations of the budget (MERLIN completes 54–55 generations instead of 59). The same step helps RAPTOR (3/4/0, 5/2/0, 1/2/0, Fig. 9b), and in the main campaign L-SHADE-SW, which adds the same step to L-SHADE, beats L-SHADE on four of seven functions at $D = 30$ without losses (1/6/0 at $D = 10$). The three algorithms differ in what happens to an accepted Solis–Wets point. In RAPTOR and L-SHADE-SW it replaces the worst member of the population and therefore takes part in the following generations. In MERLIN it only updates the incumbent: it enters neither the archive from which injected vectors are built nor the covariance update, so the search distribution never sees it. The evaluations spent on refinement improve the reported best value occasionally but do not change the course of the search, and at $D = 30$, where every generation counts, the net effect is negative. This explanation is consistent with all three comparisons, although it has not been tested directly by feeding the accepted point back into MERLIN.

The ablation also bears on the Grasshopper results. As explained in Section 3.5, MERLIN ran without its Solis–Wets step in the travelling-salesman and shell layers. In those layers it therefore used the configuration that this ablation identifies as the better one for the larger problems.

## 6.2 RAPTOR

The differential-evolution injection is the most useful addition to the swarm (4/3/0 and 5/2/0 on the rotated functions), followed by the Solis–Wets step discussed above, whose effect at D = 30 is large: without it the median grows by about one order of magnitude on four of the seven functions. On the building problem the injection is again neutral or slightly harmful (0/2/1). The Lévy relocation of stagnating particles has no measurable effect at this budget (1/5/1, 0/7/0, 0/3/0): with 60 generations, few particles stagnate long enough to trigger it, and its contribution would have to be assessed on longer runs. The velocity limit matters little in the range tested. A tighter limit of 0.2 is not worse than the default of 0.5 and is slightly better at D = 30 (0/5/2), while a limit of 1.0 is slightly worse (1/6/0, 2/5/0). The default of 0.5 was chosen on the travelling-salesman layer, where integer sliders need larger moves to change a tour; on continuous problems a tighter limit is preferable.

## 6.3 Budget-aware scheduling

Figure 2b summarizes the effect of the delayed population reduction. Switching the delay off makes all three L-SHADE variants worse: over 51 comparisons (three algorithms, seventeen problems) the schedule wins 40, ties 11 and loses none. The effect is strongest at D = 30, where the delay wins on all seven functions for L-SHADE and L-SHADE-RSP and on six for L-SHADE-SW. On the building problem it wins on four of nine comparisons and ties the rest. The schedule costs nothing at run time and leaves the algorithm unchanged at large budgets, so it can be adopted without trade-off whenever the budget is known in advance.

The analogous change to Harris hawks optimization has the opposite effect. Stretching the horizon of the escaping energy makes HHO worse on ten of seventeen problems and better on none (0/3/4, 0/2/5, 0/2/1). In L-SHADE, delaying the reduction keeps more candidates without suppressing exploitation, because every individual still competes with its own trial vector in each generation. In HHO the energy decides directly between exploration and exploitation moves, and a longer horizon keeps the hawks exploring during most of a short run, so that the phase in which they converge on the prey is cut short. A rule that works for one family does not transfer to another just because both are tied to the budget, and the HHO rescaling is not retained as part of the contribution.

## 6.4 Excluding inert inputs on the shell

On the shell the solvers of SilverHawk searched the 85 effective variables, whereas the external tools received all 100 sliders, 15 of which are overwritten by the supports (Section 4.5). To check that this difference does not explain the gap to the external tools, CMA-ES and MERLIN were run five more times for 15 minutes with all 100 sliders as variables, under otherwise identical conditions. The exclusion makes no measurable difference. With 100 variables the median of CMA-ES is 0.690 cm (0.697 cm with 85) and that of MERLIN 0.612 cm (0.696 cm), with p = 1.0 in both cases, and the number of evaluations per run is unchanged (2,214–2,458). Both solvers remain below every run of the five external tools. At this budget, fifteen inert dimensions cost neither method a measurable amount of progress, and the advantage over the external tools reported in Section 5.6 does not come from the smaller search space.

## 6.5 Summary

Three of the components evaluated earn their evaluations under a hard cap: the differential-evolution injection in both hybrids on rotated landscapes, the Solis–Wets step where its result re-enters the population (RAPTOR, L-SHADE-SW), and the delayed population reduction in the L-SHADE family. Three do not: the Solis–Wets step in MERLIN, the Lévy relocation in RAPTOR at this budget, and the horizon rescaling of HHO. The injection fraction in MERLIN and the velocity limit in RAPTOR have defaults that are not optimal at every dimension. Excluding the inert inputs of the shell has no measurable effect. None of the components helps on the building problem, where the differences between variants are small and the differential-evolution solvers lead regardless (Section 5.4).

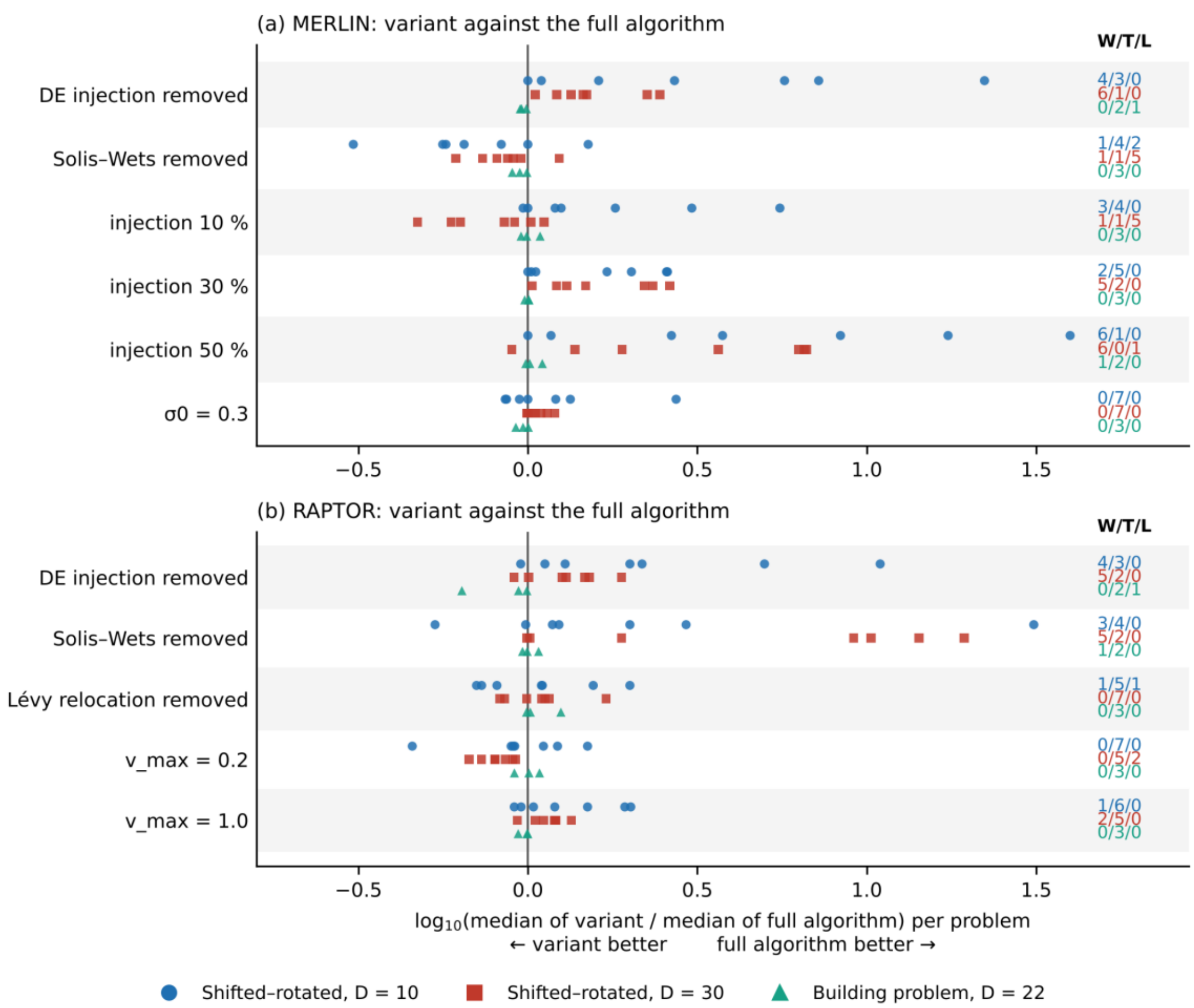

**Figure 9.** Ablation of MERLIN (a) and RAPTOR (b) under a hard cap of 1,200 evaluations, 30 runs. Each marker is one problem: the base-10 logarithm of the ratio between the median of the variant and that of the full algorithm (positive: the full algorithm is better). Markers

beyond the axis limits are clipped. Right: wins, ties and losses of the full algorithm (Mann–Whitney, $\alpha = 0.05$) on the seven shifted–rotated functions at $D = 10$ and $D = 30$ and on the three objectives of the building problem.

# 7. Discussion

## 7.1 Choosing a solver for a design problem

The results do not identify a best solver, and the paper does not claim one. They show instead that the ranking is decided by a small number of properties that a designer can usually state before the first evaluation: whether the variables are continuous and interact, whether some of them are binary or discrete, how many there are, and when the run will be stopped. Figure 10 condenses Sections 5 and 6 into a map organized by these properties.

On continuous problems with interacting variables, represented here by the shifted–rotated functions, the methods that learn a search distribution lead. MERLIN is first at both dimensions and at every checkpoint except the first quarter at $D = 10$, where PSO-IW is marginally ahead, and CMA-ES is second at the end of the run. The travelling-salesman instances, although combinatorial, behave in the same way once encoded through sliders: CMA-ES and MERLIN are the two best solvers of the roster and cannot be separated, while the difference-vector methods fall to the bottom half. The plateaus of the branch encoding reward a method that adapts the scale and orientation of its sampling, and penalize one that moves along differences between members of the population. The CMA-ES implementation of SilverHawk deserves explicit mention. It is the most consistent of all solvers outside the building problem: third on the rotated functions at $D = 10$ and second at $D = 30$, first on the travelling-salesman instances, level with MERLIN at the top on the shell, and never significantly worse than any external tool, including the CMA-ES of Opossum, which it beats on the 15-city instance and on the shell (Fig. 8). Where MERLIN is not needed or its gain is small, CMA-ES is the safest single choice.

On the building problem the order is reversed. The differential-evolution solvers, which were mid-table on the rotated functions, lead at the end of the run, and MERLIN and CMA-ES fall behind. This was attributed in Section 5.4 to the binary variables and to the dependence of the objectives on few variables. Both properties are common in building-performance models, where design options such as the presence of shading, the glazing type or the construction system are switches rather than dimensions. The free-form shell is closer to the first group: the variables are continuous and act jointly on the deformation of the surface, and MERLIN shares the first position with CMA-ES, ahead of the DE solvers.

Three solvers should be avoided in this regime. HHO is last or nearly last on the synthetic, travelling-salesman and building problems and in the lower half on the shell. PFO is in the lower part of the ranking on every problem and on the shell collapses within a few minutes. PSO-IW, competitive on the small rotated functions, is next to last on the building problem. The portfolio selector included in SilverHawk for reference (an earlier build in the travelling-salesman layer) is never last on any problem. It is second on the 10-variable rotated functions, fourth of fourteen on the shell and sixth of sixteen on the 20-city instance, ahead of all external tools except the two Opossum solvers, but it falls to the lower part of the ranking on the 30-variable rotated functions and on the 15-city instance. Its mechanism is outside the scope of this paper.

The external tools appear in the last column of Figure 10, and Section 5.6 compares them with each solver of SilverHawk. The practical conclusion is simple. A designer who replaces Galapagos, SilverEye or Tunny with CMA-ES or MERLIN obtains a result that is better or statistically equivalent on every problem run in Grasshopper, and never significantly worse (Fig. 8). On the structural problem the gain is large: the median displacement after 15 minutes falls from 2.79 cm with Galapagos, and 1.34 cm with the best external tool, to 0.70 cm with MERLIN and with CMA-ES. The only external tool that matches the best solvers is the surrogate-based Opossum-RBFOpt on small combinatorial instances, at a run time 3 to 25 times longer and with a gap that grows quickly with the instance size. These comparisons use the default settings of each tool, as a designer would, and do not measure what each tool could reach with tuning.

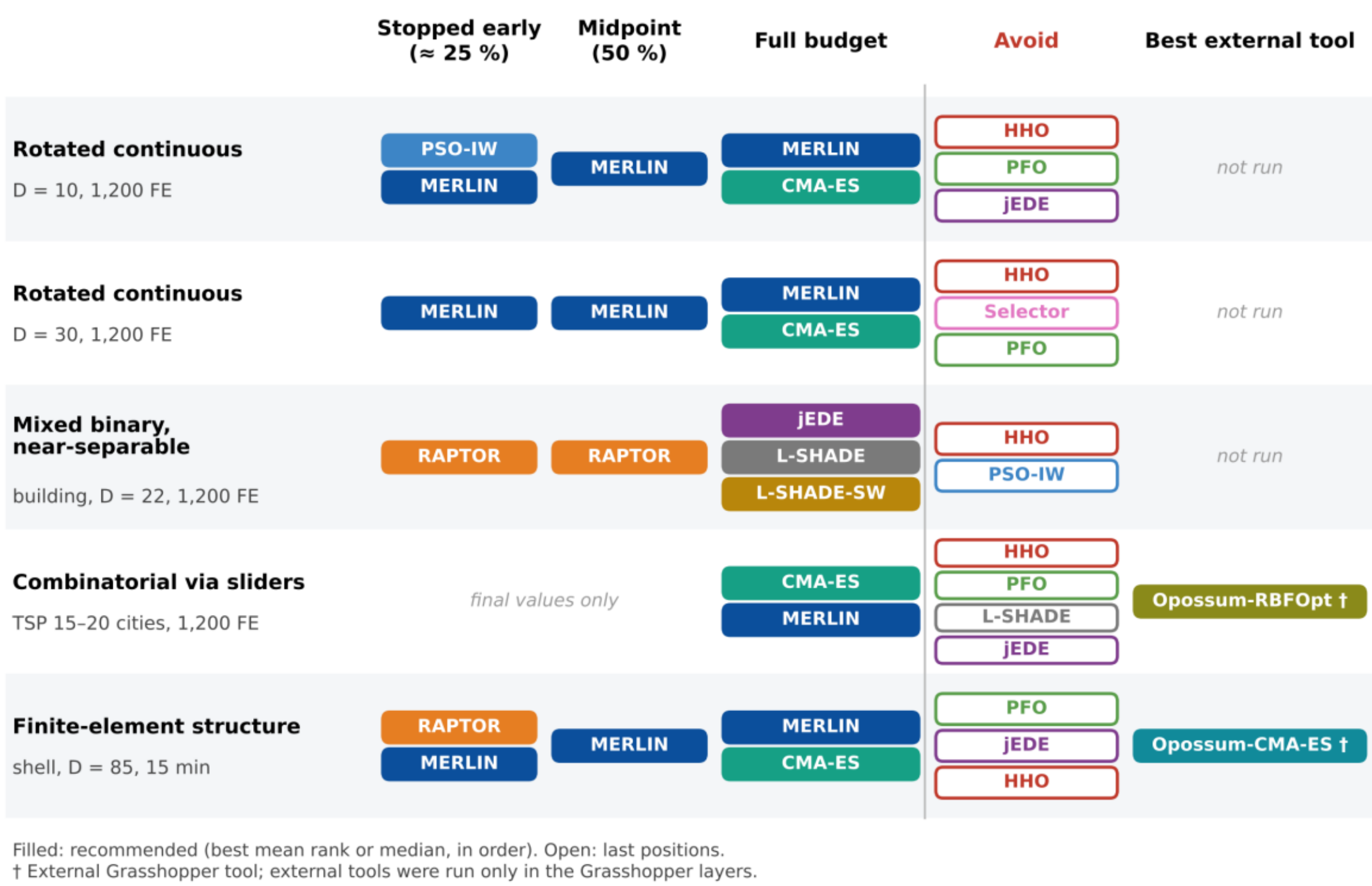

**Figure 10.** Solver-selection map for the ADO regime, derived from Tables 3–6 and Figures 5, 6 and 8. Rows: problem classes of this study. Columns: recommended solvers when the run is stopped after about a quarter of the budget, at the midpoint and at the end; solvers in the last positions; best external Grasshopper tool. Filled boxes are listed by mean rank (synthetic and building problems) or median (travelling-salesman and shell problems). External tools were run only in the Grasshopper layers; the travelling-salesman runs record final values only.

## 7.2 The centre bias and the evidence base of design studies

The reversal of HHO between centred and shifted–rotated functions (Section 5.1) has a consequence beyond this comparison. Synthetic functions with the optimum at the origin, or at the centre of the box, remain the usual sanity check in studies that introduce or select an optimizer for building design, and HHO and other recent metaheuristics are often recommended on that basis. On the centred suite of this study at D = 30, HHO would have been the first choice. On the same functions shifted and rotated, it is last, and on the three engineering problems it ends in the lower half of the ranking. A test suite that rewards convergence to the centre of the box therefore cannot support a recommendation for design problems, whose optima have no reason to lie at the centre of the parameter ranges chosen by the designer. Shifting and rotating the functions costs nothing and should be the minimum standard for synthetic comparisons in this field (Kůdela, 2022).

## 7.3 The budget horizon and the cost of time

Figure 5 shows that the ranking changes during a run, and Figure 10 makes the consequence explicit: the recommended solver depends on when the designer intends to stop. On the building problem RAPTOR is the best choice for the first half of the budget and the DE solvers for the second. On the shell, RAPTOR leads after a quarter of the time and MERLIN from the midpoint to the end. This matters in design practice, where runs are often interrupted to inspect intermediate designs, or stopped when the deadline arrives rather than when the search has converged.

The budget-aware schedule addresses the same issue from the side of the algorithm. When the budget is known at the start, the L-SHADE variants can delay population reduction accordingly, and they become better at no cost in 40 of 51 comparisons (Section 6.3). The negative result for HHO shows that this kind of adaptation must follow the mechanism of each algorithm and cannot be applied as a generic rescaling of time. Both observations also have a limit that should be stated. The anytime ranks in Figure 5 are measured within runs planned for the full budget, so they do not predict what a budget-aware solver would achieve if it were given a quarter of the budget from the start. When the run is stopped by time rather than by an evaluation count, as on the shell, the budget in evaluations is not known in advance, and the schedule cannot be used without an estimate of it (Section 3.5).

The shell adds a second distinction, between counting evaluations and counting time. A designer waits for wall-clock time and observes the best design found in that time; the number of evaluations is a consequence. On the shell, where one analysis takes 0.4 s, the surrogate-based Opossum-RBFOpt spends about 1.75 s per evaluation on fitting and searching its model, completes 418 evaluations in 15 minutes and ends at 1.90 cm, against 0.70 cm for MERLIN and CMA-ES. Counted by evaluations, the same runs say the opposite: after 418 evaluations RBFOpt is not worse than any solver of SilverHawk and better than CMA-ES and L-SHADE-RSP (Section 5.5). The two readings meet at a crossover that depends on the cost of one evaluation. With the overhead and the evaluation counts measured here, a model-based method that reached the median of MERLIN in $\nu$ times fewer evaluations would break even when the analysis takes about $1.75/(\nu - 1)$ s; the present data do not tell where $\nu$ lies at larger evaluation counts, because RBFOpt never reaches them in this time. For analyses that take minutes, as in building-energy simulation or non-linear structural analysis, the overhead is negligible and efficiency per evaluation decides, which is the regime where model-based methods have been found strong (Wortmann, 2017). For fast parametric models such as the shell, the overhead dominates and population-based solvers with negligible internal cost are preferable. Benchmarks in this field should report both axes and the cost per evaluation of each tool (Section 4.5); a ranking by evaluations alone does not tell a designer which tool to run.

### 7.4 What the hybrid components contribute

The ablations show that hybridization does not remove the dependence on the problem class but shifts the performance within it. MERLIN improves on CMA-ES wherever the covariance model of CMA-ES is appropriate, and the improvement comes from the differential-evolution injection, not from the smaller initial step size. Where CMA-ES is weak, on the building problem, the injection does not rescue it. The same holds for RAPTOR: without the injection or the Solis–Wets step it loses on most rotated functions (Section 6.2), and with both it is third at D = 30, well ahead of PSO-IW and PFO, but it does not reach the methods whose search model suits these functions better.

The Solis–Wets results suggest a design rule for components added to a population-based method under a small budget. The same step helps when its result replaces a member of the population, as in RAPTOR and L-SHADE-SW, and hurts when it only updates the incumbent, as in MERLIN. Under a hard cap a component that consumes evaluations must change the state of the search, not only the reported best value, to repay them. This reading is consistent with the three comparisons but has not been tested directly.

The versions of MERLIN and RAPTOR evaluated in this paper were fixed before the campaigns, and the ablations were run afterwards. The results therefore identify settings that could be improved, namely the Solis–Wets step in MERLIN and an injection fraction that does not decrease with the dimension. The algorithms are reported as evaluated rather than as they might be after the fact, so that the ablations remain an independent test of the design choices.

### 7.5 Limitations

The study has several limitations. All console experiments use a single budget of 1,200 evaluations. Smaller and larger budgets are represented only by the anytime ranks, with the limitation discussed in Section 7.3. The building problem is a single instance evaluated through a surrogate, and its conclusions concern this type of model rather than building simulation in general. The travelling-salesman instances were run ten times per solver and Tunny three times, which limits the power of the tests; several differences that are large in median are not significant.

The shell results rest on five runs per solver, the largest number compatible with a main campaign of about 17 hours for fourteen solvers on one machine. With five runs the Mann–Whitney test cannot reach the thresholds of a correction over all pairs, so the corrected tests were restricted to a family of six comparisons, defined during the campaign rather than registered in advance (Section 4.7), and report the others as descriptive. Differences such as MERLIN against L-SHADE ($p = 0.056$) or the ordering within the group between 0.70 and 0.90 cm should not be read as established. The runs of SilverHawk and those of Galapagos and of the Opossum 3.0.1 solvers were made five months apart with different builds; definition, machine and throughput per evaluation were the same (Section 4.5), but the comparison is not simultaneous. TPE and MADS were run in the same week as SilverHawk, with a preset number of evaluations, and their logs were cut at 900 s. In the Grasshopper layers, the L-SHADE variants ran with a constant population and MERLIN without its Solis–Wets step (Section 3.5), so the shell and travelling-salesman results characterize those configurations. PFO stops by stagnation on the shell and is not time-fair.

External tools were run only inside Grasshopper, with default settings and their own stopping criteria. Independent implementations of the same algorithm in different Grasshopper tools can return different results on one black box (Ambrosini, 2026), so the results characterize the tools as distributed rather than their algorithms in general. For Opossum-RBFOpt the time axis in Figure 5c is approximated from the evaluation count. All experiments were run on a single workstation, so absolute times will differ on other hardware, although the ratios between solvers in the same layer should not. Finally, the paper considers single-objective problems only; multi-objective search under the same budgets raises different questions and is not addressed here.

## 8. Conclusion

This paper studied single-objective optimization in the regime that architectural and structural design actually imposes: tens of variables, about a thousand evaluations of an expensive model, and runs that are often stopped by time or by the designer rather than by convergence. This paper proposed two hybrid metaheuristics for this regime, MERLIN and RAPTOR, and a budget-aware schedule for the population reduction of L-SHADE, and evaluated them with seven established algorithms and the optimizers of four Grasshopper tools under a protocol that counts every evaluation, removes the centre bias of the synthetic suite and gives equal wall-clock time on the finite-element problem.

Four results stand out. First, injecting differential-evolution trial vectors into the sampling of CMA-ES makes MERLIN the best solver on the shifted–rotated functions at both dimensions, and the ablation attributes the margin over CMA-ES to the injection. Second, no solver is best on every problem: covariance-based methods lead on continuous problems with interacting variables and on the slider-encoded travelling-salesman instances, differential evolution leads on the building problem with binary choices, and the order also changes with the point at which the run is stopped. Third, against the tools that designers use today, the two best solvers of the plug-in, MERLIN and its implementation of CMA-ES, are never significantly worse and are better than the genetic and particle-swarm tools on every problem run in Grasshopper; on the free-form shell they end 48 % below the best external tool in the same 15 minutes. A surrogate-based tool that is competitive per evaluation loses on this problem because its overhead exceeds the cost of the analysis, so benchmarks for designers should report wall-clock time as well as evaluations. Fourth, two methodological points apply beyond this study: synthetic functions centred in the search box can reverse a ranking, as they do for Harris hawks optimization, and under a hard cap a component must change the state of the search, not only the reported best value, to repay its evaluations.

For practice, these results translate into a short rule (Fig. 10). On continuous problems with interacting variables, use MERLIN or CMA-ES. On problems with binary or discrete switches, use RAPTOR if the run will be short and an adaptive differential-evolution method if it will use the full budget. Do not choose a solver on the basis of centred test functions.

The ablations also indicate where the proposed algorithms can be improved. The injection fraction of MERLIN should decrease with the dimension, and the result of its Solis–Wets step should be fed back into the search distribution or the archive rather than only into the incumbent. A budget-aware schedule for time-limited runs requires an estimate of the evaluation budget from the elapsed time, which the parametric host can provide during the run. Warm starts that spend part of a short run on choosing a starting point, and the extension of the same protocol to multi-objective problems, are further directions.

## Data availability

The run data, the problem instances (city coordinates, shift vectors and rotation matrices) and the analysis scripts are available from the author on reasonable request and will be deposited on Zenodo upon journal publication.

## Declaration of competing interest

The author develops the SilverHawk plug-in evaluated in this paper, which will be distributed free of charge. The author declares no other competing interests.

## Funding

This research did not receive any specific grant from funding agencies in the public, commercial or not-for-profit sectors.

## Declaration of generative AI and AI-assisted technologies in the manuscript preparation process

During the preparation of this work the author used Claude (Anthropic; Claude Sonnet 4.6 and Opus 4.6 in May–June 2026, Claude 5.5 in October 2026) to assist with code for data analysis and figures, the checking of numbers against the raw data, and language editing. After using this tool, the author reviewed and edited the content as needed and takes full responsibility for the content of the published article.